\PassOptionsToPackage{dvipsnames,table}{xcolor}

\documentclass{article} %
\usepackage[final]{colm2025_conference} %

\usepackage{microtype}
\usepackage{hyperref}
\usepackage{url}
\usepackage{booktabs}
\usepackage{multirow}
\usepackage{multicol}

\usepackage{lineno}
\usepackage{graphicx}
\usepackage{pifont}%

\newcommand{\cmark}{\ding{51}}
\newcommand{\xmark}{\ding{55}}
\newcommand{\V}{\textcolor{BrickRed}{\cmark}}
\newcommand{\X}{\textcolor{Green}{\xmark}}

\usepackage{algorithm}
\usepackage{algpseudocode}

\definecolor{darkblue}{rgb}{0, 0, 0.5}
\hypersetup{colorlinks=true, citecolor=darkblue, linkcolor=darkblue, urlcolor=darkblue}

\title{Evaluating Large Language Models as Expert Annotators}

\author{
    Yu-Min Tseng$^{\alpha\beta\dagger}$ \quad Wei-Lin Chen$^\gamma$ \quad Chung-Chi Chen$^\delta$ \quad Hsin-Hsi Chen$^{\alpha\pi}$
    \\
    ~$^\alpha$National Taiwan University\quad
    ~$^\beta$Virginia Tech\quad
    ~$^\gamma$University of Virginia\quad \\
    ~$^\delta$AIST, Japan\quad
    ~$^\pi$AINTU, Taiwan
    \\
\texttt{
    ymtseng@vt.edu, 
    wlchen@virginia.edu,
    c.c.chen@acm.org,
    hhchen@ntu.edu.tw}
}

\begin{document}

\ifcolmsubmission
\linenumbers
\fi

\maketitle

\begin{abstract}
Textual data annotation, the process of labeling or tagging text with relevant information, is typically costly, time-consuming, and labor-intensive.
While large language models (LLMs) have demonstrated their potential as direct alternatives to human annotators for general domains natural language processing (NLP) tasks, their effectiveness on annotation tasks in domains requiring expert knowledge remains underexplored.
In this paper, we investigate: whether top-performing LLMs, which might be perceived as having expert-level proficiency in academic and professional benchmarks, can serve as direct alternatives to human expert annotators?
To this end, we evaluate both individual LLMs and multi-agent approaches across three highly specialized domains: finance, biomedicine, and law.
Specifically, we propose a multi-agent discussion framework to simulate a group of human annotators, where LLMs are tasked to engage in discussions by considering others’ annotations and justifications before finalizing their labels.
Additionally, we incorporate reasoning models (\textit{e.g.}, o3-mini) to enable a more comprehensive comparison.
Our empirical results reveal that:
\textit{(1)} Individual LLMs equipped with inference-time techniques (\textit{e.g.}, chain-of-thought (CoT), self-consistency) show only marginal or even negative performance gains, contrary to prior literature suggesting their broad effectiveness.
\textit{(2)} Overall, reasoning models do not demonstrate statistically significant improvements over non-reasoning models in most settings.
This suggests that extended long CoT provides relatively limited benefits for data annotation in specialized domains.
\textit{(3)} Certain model behaviors emerge in the multi-agent discussion environment.
For instance, Claude 3.7 Sonnet with thinking rarely changes its initial annotations, even when other agents provide correct annotations or valid reasoning.\footnotemark
\begingroup\def\thefootnote{$\dagger$}\footnotetext{Work was done at National Taiwan University.}\endgroup
\footnotetext{\url{https://github.com/ymntseng/llm-expert-annotators}}
\end{abstract}

\section{Introduction}

Textual Data annotation refers to the task of labeling or tagging text with relevant information~\citep{tan2024large}.
For example, adding topical keywords to social media contents.
Typically, this process is carried out by crowd-sourced workers (\textit{e.g.}, MTurkers) or specialized annotators (\textit{e.g.}, researchers), depending on the tasks, to ensure high-quality annotations.
However, the annotating procedures are often costly, time-consuming, and labor-intensive, particularly for tasks that require domain expertise.

With the rise of large language models (LLMs), a series of works have explored their potential as an attractive alternative to human annotators~\citep{ding2023gpt,zhang2023llmaaa,choi2024gpts,he2023annollm}.
Empirical results suggest that, in certain scenarios, LLMs such as ChatGPT and GPT-3.5 even outperform master-level MTurk workers, with substantially lower per-annotation cost~\citep{gilardi2023chatgpt,alizadeh2023open,bansal2023large,zhu2023can}.
However, existing studies mainly focus on general-domain NLP tasks (\textit{e.g.}, sentiment classification, word-sense disambiguation).
The extent to which LLMs as data annotators perform in domains requiring expert knowledge remains unexplored.

On the other hand, LLMs have exhibited striking performance in a variety of benchmarks, both professional and academic~\citep{jin2019pubmedqa,hendrycks2020measuring,chen2021evaluating,rein2023gpqa,achiam2023gpt}.
Leveraging the abundant domain-specific knowledge encoded in the parameters, LLMs could pass exams that require expert-level abilities~\citep{choi2021chatgpt,singhal2023large,callanan2023can,singhal2023towards,katz2024gpt}.
These findings prompt our research question: Can performant LLMs, which might be perceived as having expert-level proficiency in academic and professional benchmarks, serve as direct alternatives to human expert annotators?
We refer to this setting as \textit{LLMs-as-Expert-Annotators}.

To investigate the question, we examine LLMs on three specialized domains: finance, law, and biomedicine.
Specifically, we carefully select five existing datasets that
\textit{(i)} provide fully-detailed annotation guidelines and
\textit{(ii)} are manually labelled by domain experts.
We format the annotation task, the guideline, and unlabelled data instances as instructional inputs to the models, and evaluate their annotation results against ground truth labeled by human experts.
Toward a more comprehensive evaluation, we employ a variety of inference-time techniques that leverage additional compute to elicit the capabilities of LLMs, using both individual LLM and multi-agent (\textit{i.e.}, multiple LLMs) approaches.
Furthermore, inspired by how human annotators reach consensus, we propose a multi-agent annotation framework that allows LLMs to collaboratively generate annotations through discussion.

In sum, our main contributions includes:
\begin{itemize}
    \item We present one of the first systematic evaluations of LLMs-as-Expert-Annotators and investigate their inability to perform annotation tasks that require specialized domain knowledge.

    \item We find that, for individual LLMs
    \textit{(1)} equipping with inference-time techniques demonstrate only marginal or even negative performance gains, contrary to prior literature suggesting their broad effectiveness;
    \textit{(2)} reasoning models do not exhibit statistically significant improvements over non-reasoning models in most settings.

    \item We propose a multi-agent discussion framework that enables multiple LLMs to reach stronger consensus, leading to improved performance over individual LLMs.
    In addition, we conduct a fine-grained analysis of the results and identify specific model behaviors that emerge during the multi-agent discussion process.

\end{itemize}

Given the high-stakes nature of expert-level annotation in fields such as medicine and finance, our findings highlight a notable gap between current LLMs and human experts, underscoring the need for further advancements before LLMs can be reliably deployed as expert annotators.

\section{Experimental Setup}
\subsection{Datasets}
We evaluate five datasets across three specialized domains: finance, law, and biomedicine. (Task descriptions, dataset statistics, and annotation guidelines are provided in Appendix~\ref{appendix-dataset} and~\ref{subsec:appendix-guideline}.)
All datasets are multiple-choice tasks.
Due to limited resources, we sample 200 instances per dataset, totaling 1000 instances.
To ensure data quality and a fair comparison, we have checked these datasets \textit{(1)} provide fully documented annotation guidelines and \textit{(2)} explicitly state that annotation were labeled by human experts and reach consensus.

\subsection{Models}
We experiment with 6 of the most performant, publicly-available language models, including 4 non-reasoning models and 2 reasoning models.
The non-reasoning models are \texttt{Gemini-1.5-Pro}~\citep{reid2024gemini}, \texttt{Gemini-2.0-Flash}~\citep{google2024gemini2.0}, \texttt{Claude-3-Opus}~\citep{anthropic2024claude}, and \texttt{GPT-4o}~\citep{openai2024gpt4o}.
The reasoning models are \texttt{Claude-3.7-Sonnet}~\citep{anthropic2025claude} with thinking and \texttt{o3-mini}~\citep{openai2025o3mini} with medium reasoning effort.
For models with a temperature parameter, we set it to 0.0 unless otherwise specified in the method settings.

\subsection{Evaluation}

We assess LLMs-as-expert-annotators by comparing their accuracy against ground-truth labels provided by human expert annotators.

Unlike prior studies that evaluate LLMs on general-domain datasets, we do not compare their performance with crowdworker or non-expert human annotations.
Our investigation targets datasets that require specialized domain expertise, which crowdworkers might not be able to provide satisfactory annotations. 
Further recruiting new human annotators for this study could result in an unfair comparison, as the original dataset annotations were produced by highly selective, qualified experts.
Furthermore, using gold-standard annotations provides a more suitable and reproducible test bed for future works to compare the results directly.

\section{Individual LLMs as Expert Annotators}
In this section, we adopt vanilla prompting along with three inference-time techniques: CoT, self-refine, and self-consistency.
By leveraging additional inference-time compute, we explore whether individual LLMs can serve as a direct alternative to expert data annotators.

We employ a uniform prompt template that is easily generalizable across all models and tasks.
This standardization of prompt phrasing ensures that the only sources of variation in our results are: \textit{(i)} the annotation guideline and \textit{(ii)} the instance to be labeled.
We provide all prompt templates in Appendix~\ref{subsec:appendix-prompt}.

\subsection{Methods}

\paragraph{Vanilla}
The vanilla method refers to standard direct-answer prompting, where instructional input consists of the annotation task, guideline, and the instance are given to the LLMs.
LLMs are tasked to conduct annotation as a domain expert of relevant fields.
The vanilla prompt also serves as the base of other sophisticated approaches (described below).

\paragraph{CoT} 
CoT improves LLMs' complex reasoning ability significantly~\citep{wei2022chain}.
Specifically, we employ zero-shot CoT~\citep{kojima2022large}, where a trigger phrase ``\textit{Let’s think step by
step}'' augments the prompt to elicit reasoning chain from LLMs and leads to a more accurate answer.

\paragraph{Self-Consistency}
Self-consistency~\citep{wang2022self} improves upon CoT via a sample-and-marginalize decoding procedure, which selects the most consistent answer rather than the greedily decoded one.
Concretely, we sample 5 diverse reasoning paths with temperature 0.7, and take the majority vote to determine the final answer.

\paragraph{Self-Refine}
Self-refine~\citep{madaan2024self} method includes three steps: generate, review, and refine.
An LLM first generates an initial answer (\textit{i.e.}, draft).
Then, the model review its draft and provide feedback.
Lastly, the LLM refine the draft by incorporating  its feedback, and outputs an improved answer.
The same LLM is used in all steps.

\subsection{Inference-Time Techniques Could Undermine LLMs-as-Expert-Annotators}
Our results in Table~\ref{tab:individual-result} suggest that models struggle to effectively and consistently leverage inference-time techniques, often experiencing performance declines.

Across all models, the application of CoT generally leads to lower accuracy. 
For instance, Claude 3 Opus exhibits an average accuracy drop of 1.6\%. 
Even when minor improvements are observed in specific datasets, they are inconsistent and fail to establish a reliable trend of enhancement.
Similarly, self-refine and self-consistency show unstable effects.
While some cases exhibit slight gains, most results  reflect a negative impact, such as GPT-4o experiences an average 1.4\% decrease when equipped with self-refine method.

This overall performance decline suggests that these inference-time techniques, which have demonstrated significant performance gains in prior literature, may not be well-suited for the LLMs-as-expert-annotators setting.
We speculate that inference-time methods may fail to consistently enhance performance due to fundamental limitations in models' ability to understand complex domain-specific contexts.
Specifically, models might not accurately interpret specialized annotation guidelines and input instances, thereby failing to capitalize on the additional inference-time compute, or even degrading performance due to misinterpretation.
Therefore, instruction-tuned models appear to struggle with applying these strategies effectively, highlighting a critical limitation in their ability to perform data annotation tasks in specialized domains.

\begin{table}[t]
    \small
    \resizebox{\textwidth}{!}{
  \centering
    \begin{tabular}{lrrrrrrrrr}
    \toprule
    \multicolumn{1}{c}{\multirow{2}[4]{*}{\textbf{Model / Method}}} & \multicolumn{2}{c}{\textbf{Finance}} &       & \multicolumn{2}{c}{\textbf{Law}} &       & \multicolumn{1}{l}{\textbf{Biomedicine}} &       & \multicolumn{1}{c}{\multirow{2}[4]{*}{\textbf{Avg.}}} \\
\cmidrule{2-3}\cmidrule{5-6}\cmidrule{8-8}          & \multicolumn{1}{c}{\textbf{REFinD}} & \multicolumn{1}{c}{\textbf{FOMC}} &       & \multicolumn{1}{c}{\textbf{CUAD}} & \multicolumn{1}{c}{\textbf{FoDS}} &       & \multicolumn{1}{c}{\textbf{CODA-19}} &       &  \\
    \midrule
    
    \textit{Claude 3 Opus} & 64.0  & 63.0  &       & 83.5  & 47.0  &       & 63.5  &       & 64.2 \\
    
    \textit{w/} CoT & 62.0 (\textcolor{purple}{$\downarrow$2.0})  & 64.5 (\textcolor{teal}{$\uparrow$1.5}) &       & 80.5 (\textcolor{purple}{$\downarrow$3.0})  & 43.0 (\textcolor{purple}{$\downarrow$4.0})  &       & 63.0 (\textcolor{purple}{$\downarrow$0.5})  &       & 62.6 (\textcolor{purple}{$\downarrow$1.6}) \\
    \midrule
    
    \textit{Gemini 1.5 Pro} & 63.5  & 66.5  &       & 84.0  & 43.0  &       & 69.5  &       & 65.3 \\
    
    \textit{w/} CoT & 59.0 (\textcolor{purple}{$\downarrow$4.5})  & 68.0 (\textcolor{teal}{$\uparrow$1.5}) &       & 82.0 (\textcolor{purple}{$\downarrow$2.0}) & 37.5 (\textcolor{purple}{$\downarrow$5.5}) &       & 71.5 (\textcolor{teal}{$\uparrow$2.0}) &       & 63.6 (\textcolor{purple}{$\downarrow$1.7}) \\
    \midrule
    
    \textit{Gemini 2.0 Flash} & 59.0  & \textbf{71.5} &       & \textbf{86.5} & 47.0  &       & \textbf{79.5} &       & 68.7 \\
    
    \textit{w/} CoT & 62.5 (\textcolor{teal}{$\uparrow$3.5}) & 68.5 (\textcolor{purple}{$\downarrow$3.0}) &       & 82.5 (\textcolor{purple}{$\downarrow$4.0}) & 45.5 (\textcolor{purple}{$\downarrow$1.5}) &       & 77.5 (\textcolor{purple}{$\downarrow$2.0}) &       & 67.3 (\textcolor{purple}{$\downarrow$1.4})\\
    
    \textit{w/} self-refine & 64.5 (\textcolor{teal}{$\uparrow$5.5}) & 68.5 (\textcolor{purple}{$\downarrow$3.0}) &       & 85.0 (\textcolor{purple}{$\downarrow$1.5}) & \textbf{47.5} (\textcolor{teal}{$\uparrow$0.5}) &       & 76.5 (\textcolor{purple}{$\downarrow$3.0}) &       & 68.4 (\textcolor{purple}{$\downarrow$0.3})\\
    
    \textit{w/} self-consistency & 65.0 (\textcolor{teal}{$\uparrow$6.0})  & 70.0 (\textcolor{purple}{$\downarrow$1.5}) &       & 83.5 (\textcolor{purple}{$\downarrow$3.0}) & 46.5 (\textcolor{purple}{$\downarrow$0.5}) &       & \textbf{79.5} (\textcolor{black}{\textminus0.0}) &       & \textbf{68.9} (\textcolor{teal}{$\uparrow$0.2}) \\
    \midrule
    
    \textit{GPT-4o} & 67.5  & 68.5  &       & 84.5  & 44.5  &       & 74.0  &       & 67.8 \\
    
    \textit{w/} CoT & 67.0 (\textcolor{purple}{$\downarrow$0.5}) & 69.5 (\textcolor{teal}{$\uparrow$1.0}) &       & 84.0 (\textcolor{purple}{$\downarrow$0.5}) & 44.0 (\textcolor{purple}{$\downarrow$0.5}) &       & 72.5 (\textcolor{purple}{$\downarrow$1.5}) &       & 67.4 (\textcolor{purple}{$\downarrow$0.4}) \\
    
    \textit{w/} self-refine & 66.5 (\textcolor{purple}{$\downarrow$1.0}) & 67.0 (\textcolor{purple}{$\downarrow$1.5}) &       & 81.5 (\textcolor{purple}{$\downarrow$3.0}) & 45.0 (\textcolor{teal}{$\uparrow$0.5}) &       & 72.0 (\textcolor{purple}{$\downarrow$2.0}) &       & 66.4 (\textcolor{purple}{$\downarrow$1.4}) \\
    
    \textit{w/} self-consistency & \textbf{69.5} (\textcolor{teal}{$\uparrow$2.0}) & 69.5 (\textcolor{teal}{$\uparrow$1.0}) &       & 83.5 (\textcolor{purple}{$\downarrow$1.0}) & 46.0 (\textcolor{teal}{$\uparrow$1.5}) &       & 74.0 (\textcolor{black}{\textminus0.0}) &       & 68.5 (\textcolor{teal}{$\uparrow$0.7}) \\
    
    \bottomrule
    \end{tabular}%
    }
  \caption{Accuracy of instruction-tuned LLMs on expert annotation tasks. Text in \textbf{bold} indicates the highest accuracy for each dataset.}
  \label{tab:individual-result}%
\end{table}%

\begin{figure}
    \centering
    \includegraphics[width=1\linewidth]{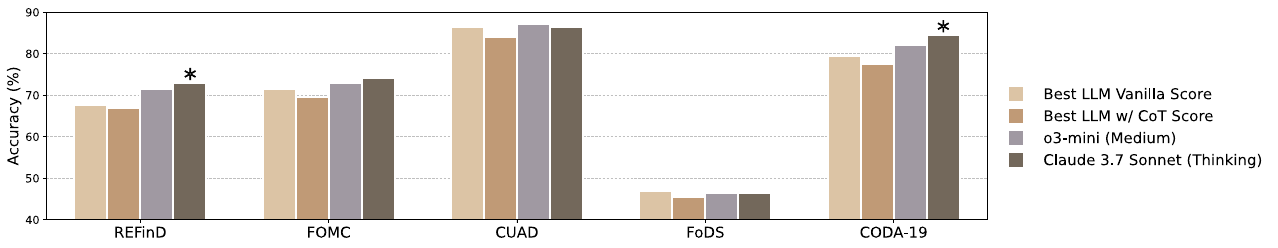}
    \caption{Accuracy comparison between reasoning models and non-reasoning models. An asterisk ($^*$) indicates the reasoning model is statistically significant with p-value~$<$~0.05 than the best non-reasoning models with CoT method.}
    \label{fig:reasoning-vs-instruction-tuned}
\end{figure}

\subsection{Reasoning Models Outperform Non-Reasoning Models Marginally}

As shown in Figure~\ref{fig:reasoning-vs-instruction-tuned}, we compare reasoning models (\textit{i.e.}, o3-mini with medium reasoning effort and Claude 3.7 Sonnet with thinking) against the best-performing non-reasoning models across five datasets.

While reasoning models tend to achieve a slightly higher accuracy, the differences are relatively small in many cases.
We apply McNemar’s test~\citep{mcnemar1947note} to assess the statistical significance of their performance differences.
The accuracy and corresponding p-values are provided in Appendix~\ref{appendix-test}.
Across four comparison settings (\textit{i.e.}, each of the two reasoning models versus each of the two non-reasoning models), statistical significance is observed in only one setting -- when comparing Claude 3.7 Sonnet with thinking against the best non-reasoning models with CoT -- and in two out of five datasets.
These results suggest that, despite their enhanced long CoT inference capabilities, current reasoning models do not yet offer a substantial advantage over non-reasoning models in the LLM-as-expert-annotators setting.

\section{Multi-Agent Discussion Framework}

The multi-agent framework, where multiple LLMs communicate with each other to solve tasks in a collaborative manner, has become a prevalent research direction~\citep{liang2023encouraging,du2023improving,chen2023reconcile,tseng2024two}.
This approach leverages the collective power of multiple models, enabling them to exchange insights, verify conclusions, and reduce individual biases, ultimately enhancing task performance and decision quality.

In the context of data annotation, a common challenge is the disagreement among multiple annotators, where differing interpretations lead to inconsistent labels.
In human annotation workflows, such discrepancies are often resolved through peer discussions, where annotators deliberate over ambiguous cases to reach a consensus.
Inspired by this collaborative resolution process, we design a multi-agent annotation framework that incorporates a discussion mechanism.
By enabling LLM agents to engage in communication, the framework simulates the consensus-building process of human annotators to assess whether this approach results in more accurate and reliable annotations.

\subsection{Proposed Framework}

\begin{algorithm}
\caption{Multi-Agent Discussion Framework Algorithm}
\label{alg:peer_discussion}
\begin{algorithmic}[1]
    \Require Set of LLMs $\mathcal{M} = \{M_1, M_2, ..., M_n\}$  
    \Require Annotation task $T$, Annotation guideline $G$, Instance $I$  
    \Require Maximum discussion rounds $R_{\max}$
    \Ensure Final annotation $\hat{y}$ 
    
    \State \textbf{Initialization:} Set discussion round counter $r \gets 0$
    
    \State \textcolor{purple}{\textbf{Step 1: Generate Initial Annotation}}
    \For{each $M_i \in \mathcal{M}$}
        \State $\hat{y}_i \gets M_i(T, G, I)$
    \EndFor
    
    \While{$r < R_{\max}$}
        \State \textcolor{purple}{\textbf{Step 2: Check Consensus}}
        
        \If {all annotations $\hat{y}_1, \hat{y}_2, ..., \hat{y}_n$ are identical}
            \State \Return Final annotation $\hat{y}=\hat{y}_1= \hat{y}_2= ... = \hat{y}_n$
        \Else
            \State \textcolor{purple}{\textbf{Step 3: Discuss and Re-Annotate}}
            \State Compile all $\hat{y}_i$ and reasoning into Discussion History $D_r$
            \For{each $M_i \in \mathcal{M}$}
                \State Generate revised annotation: $\hat{y}_i^{(r+1)} \gets M_i(T, G, I, D_r)$
            \EndFor
            \State $r \gets r + 1$
        \EndIf
    \EndWhile
    \State \textcolor{purple}{\textbf{Step 4: Majority Vote (if no consensus reached)}}
    \State $\hat{y} = Majority~Vote(\hat{y}_1, \hat{y}_2, ..., \hat{y}_n)$
    \State \Return $\hat{y}$
\end{algorithmic}
\end{algorithm}

Our proposed multi-agent discussion framework involves four steps: (\textit{1}) Generate initial annotations, (\textit{2}) Check consensus, (\textit{3}) Discuss and re-annotate, and \textit{(4)} Majority vote if no consensus reach, as illustrated in pseudo  algorithm~\ref{alg:peer_discussion}.
Initially, each agent generates its own annotation through CoT prompting given the same annotation task, guideline, and instance.
Next, we check for consensus (\textit{i.e.}, if all annotations are the same labels).
If consensus is achieved, the instance is successfully annotated and the process completes.
If not, we compile all agents' reasoning and labels into a ``\textit{Discussion History}''.
Agents then re-annotate using the same input and discussion history.
This check-discuss-re-annotate cycle continues until consensus is achieved or the maximum number of discussion round is reached. 
We provide the prompt templates of our proposed framework in Appendix~\ref{subsec:appendix-prompt}.

In our experiment, we set the maximum number of discussion rounds to 2 based on empirical observations and practical considerations:
\textit{(i)} We find that nearly all instances reach consensus within 2 rounds. In fact, fewer than 10 instances fail to reach agreement by the end of the second round.
\textit{(ii)} For the rare cases that do not reach consensus after 2 rounds, we observe a common pattern -- all three agents resist changing their annotations in at least one round, resulting in no progress.

Through the average performance of individual reasoning models and non-reasoning models, we select three representative model pairings for the multi-agent framework:
\begin{itemize}
    \item Claude 3.7 Sonnet with thinking, o3-mini, GPT-4o (2 reasoning models)
    \item Claude 3.7 Sonnet with thinking, GPT-4o, Gemini 2.0 Flash (1 reasoning models)
    \item GPT-4o, Gemini 2.0 Flash, Gemini 1.5 Pro (0 reasoning models)
\end{itemize}
To enhance the majority vote accuracy, we set the multi-agent group size (i.e., number of LLMs) to 3, ensuring a more robust and reliable consensus outcome.

\subsection{Multi-Agent Discussion Leads to Better Agreement and Performance}

\begin{figure}[!ht]
    \centering
    
    \begin{minipage}{1\textwidth}
        \centering
        \includegraphics[width=\linewidth]{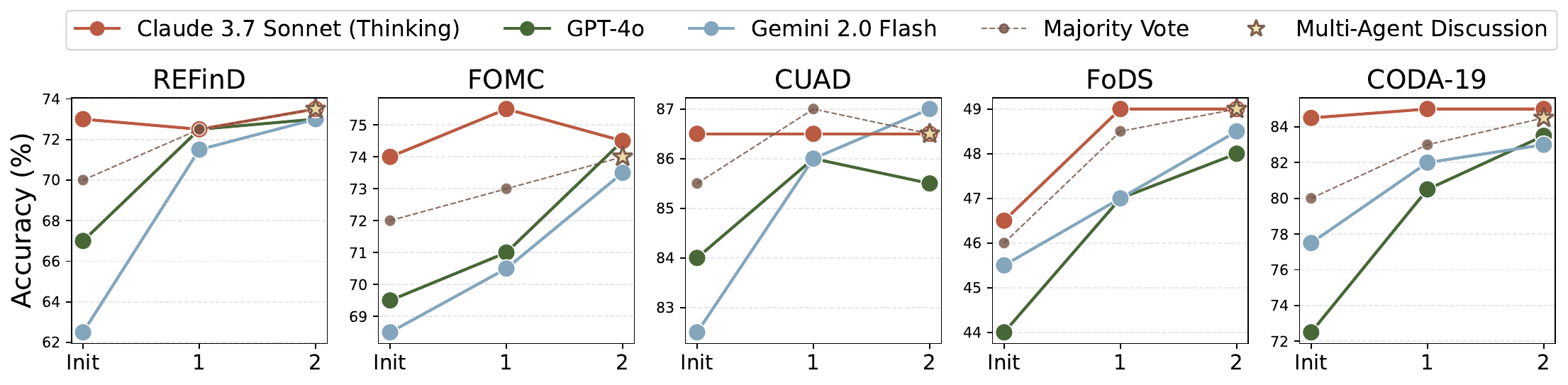}
    \end{minipage}
    \hfill
    \begin{minipage}{1\textwidth}
        \centering
        \includegraphics[width=\linewidth]{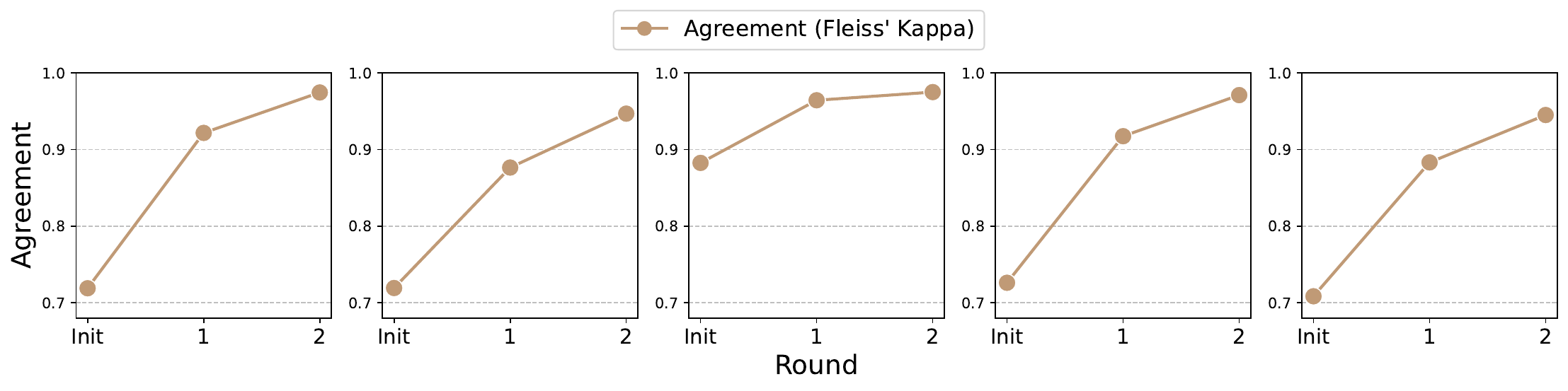}
    \end{minipage}

    \caption{Accuracy and agreement improvements through multi-agent discussion across five datasets (1 reasoning model setting). The top row shows the accuracy progression of each agent and the whole discussion framework. The bottom row displays the Fleiss’ Kappa agreement scores, indicating improved consensus among the models.}
    \label{fig:multi-llm-agreement}
\end{figure}

The multi-agent discussion framework demonstrates its effectiveness by enhancing both accuracy and inter-annotator agreement across all five datasets, as shown in Figure~\ref{fig:multi-llm-agreement}.

For performance (top row of the figure), the accuracy of both the discussion framework and each individual agent increases as the discussion rounds progress, demonstrating that collaborative interactions among LLMs lead to more accurate annotations.
As the agents iteratively refine their reasoning through communication, the overall performance improves consistently.

Additionally, we calculate the Fleiss’ Kappa~\citep{fleiss1971measuring} agreement score, which measures the consistency among annotators.
As shown in the bottom row of the figure, the agreement score steadily increases with each round, indicating that multi-agent discussions not only enhance accuracy but also promote greater consensus among the models.
The final round achieves near-perfect agreement across all datasets, highlighting the robustness and reliability of the multi-agent framework in resolving ambiguous or conflicting cases.

\begin{figure}
    \centering
    \includegraphics[width=1\linewidth]{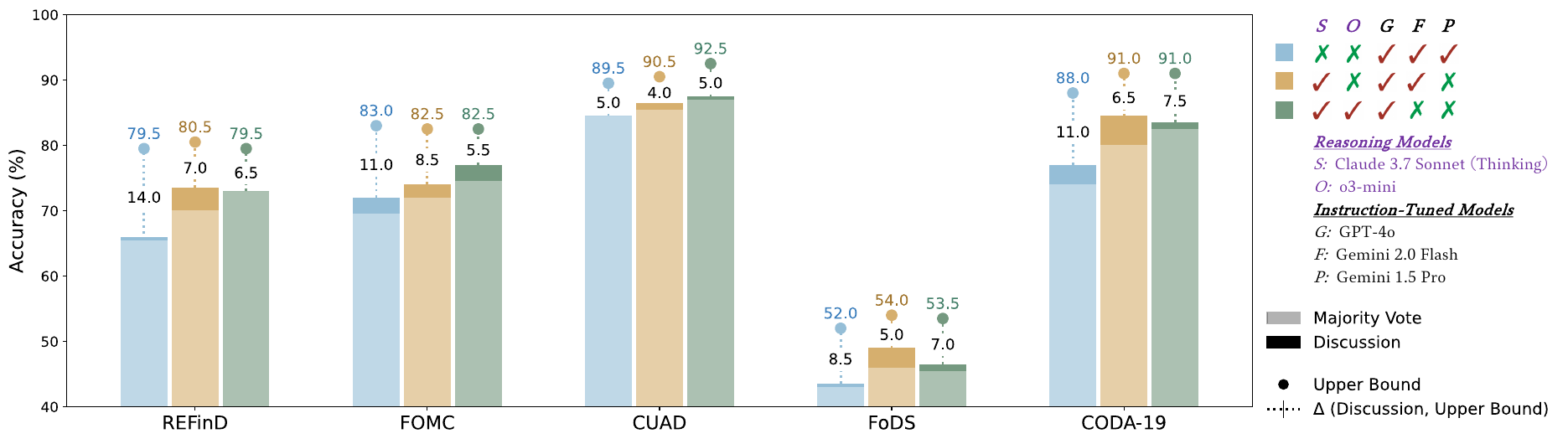}
    \caption{Comparison between multi-agent CoT MV and discussion framework for three model pairing. The dotted lines with numbers indicate the gap ($\Delta$) between the discussion framework and its upper bound.}
    \label{fig:upper-bound}
\end{figure}

\subsection{The Discussion Framework Falls Short of Its Upper Bound}

For our multi-agent discussion framework, we approximate the upper bound by assuming that agents cannot suddenly generate the correct annotation in the middle of the discussion.
The only way to arrive at a correct annotation is if it was present in at least one of the initial agent predictions, and agents are able to identify it and reach consensus through discussion.
On the other hand, if the correct annotation is absent from the initial predictions of all three agents, we treat it as an irrecoverable instance.
In these cases, regardless of the quality or depth of the discussion, the agents are unable to produce the correct annotation, as it was never part of the initial pool of responses.

As shown in Figure~\ref{fig:upper-bound}, settings containing reasoning models outperform settings containing all three non-reasoning models for multi-agent discussion framework.
While the framework consistently outperforms the CoT majority vote (MV) based on the initial annotations of the three agents, it does not always reach its upper-bound performance.
The dotted lines reveal a persistent gap ($\Delta$) between the discussion framework and its upper bound.

\subsection{Emerging Model Behaviors in the Discussion Process}

To understand why the discussion framework underperforms relative to its upper bound, we analyze the underlying model behaviors.
Since these behaviors are consistent across datasets and settings, we use the FoDS dataset with the 1 reasoning model setting as a representative example, as illustrated in Figure~\ref{fig:model-behavior}.
The figure shows the progression of annotations across two rounds (R1 and R2), visualizing how models refine or retain their predictions during the discussion process.

We observe that the reasoning model (Claude 3.7 Sonnet with thinking) rarely changes its initial annotations, regardless of their correctness.
While maintaining correct annotations is desirable, this strong self-consistency on incorrect ones  limits the potential for collaborative refinement.
By contrast, Gemini 2.0 Flash and GPT-4o exhibit greater responsiveness to peer reasoning and a higher willingness to revise annotations.
However, their revisions are not reliably targeted toward correcting only incorrect labels.
These behavioral tendencies help explain why our multi-agent framework improves over individual models, yet still falls short of the upper bound: collaborative gains are constrained by both strong self-consistency and imprecise revision behavior.

We hypothesize two explanations for the strong self-consistency observed in reasoning models:
\textit{(i)} Overconfidence.
Strong reasoning models may exhibit higher confidence in their initial annotations, making them less likely to update even when faced with correct annotations from other models.
\textit{(ii)} Greater persuasiveness.
As shown in prior work~\citep{bozdag2025must, durmus2024persuasion}, prompts that encourage logical reasoning can yield more persuasive arguments.
Consequently, reasoning models may generate outputs that are not only logically coherent but plausible, leading others to align with them, even when it's incorrect.

This observation highlights the importance of careful design in future multi-agent systems.
Whether this strong self-consistency effect is ultimately beneficial or detrimental remains an open question.
Should the effect generalize and contribute to unintended consequences, it would warrant a closer look into the role of reasoning models in collaborative LLM-as-expert-annotator systems, with consideration for appropriate safeguards.

\definecolor{darkr}{RGB}{166,113,138}
\definecolor{darkg}{RGB}{61,91,101}
\definecolor{darkb}{RGB}{160,192,215}

\definecolor{lightr}{RGB}{220,192,215}
\definecolor{lightg}{RGB}{173,190,171}
\definecolor{lightb}{RGB}{227,231,235}

\begin{figure}[!t]
    \centering
    
    \begin{minipage}{0.49\textwidth}
        \centering
        \includegraphics[width=\linewidth]{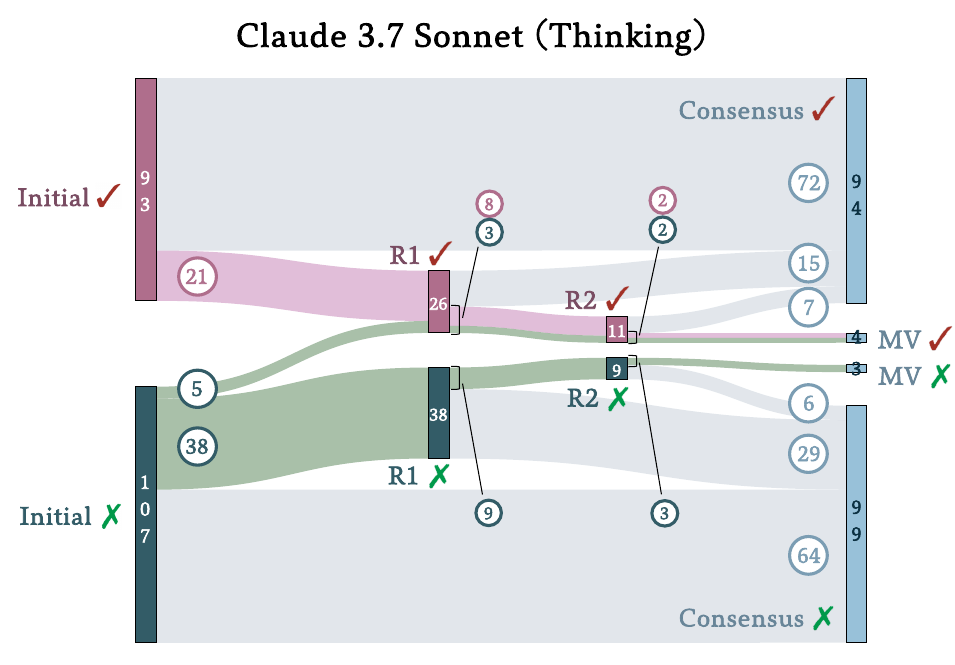}
    \end{minipage}
    \hfill
    \begin{minipage}{0.49\textwidth}
        \centering
        \includegraphics[width=\linewidth]{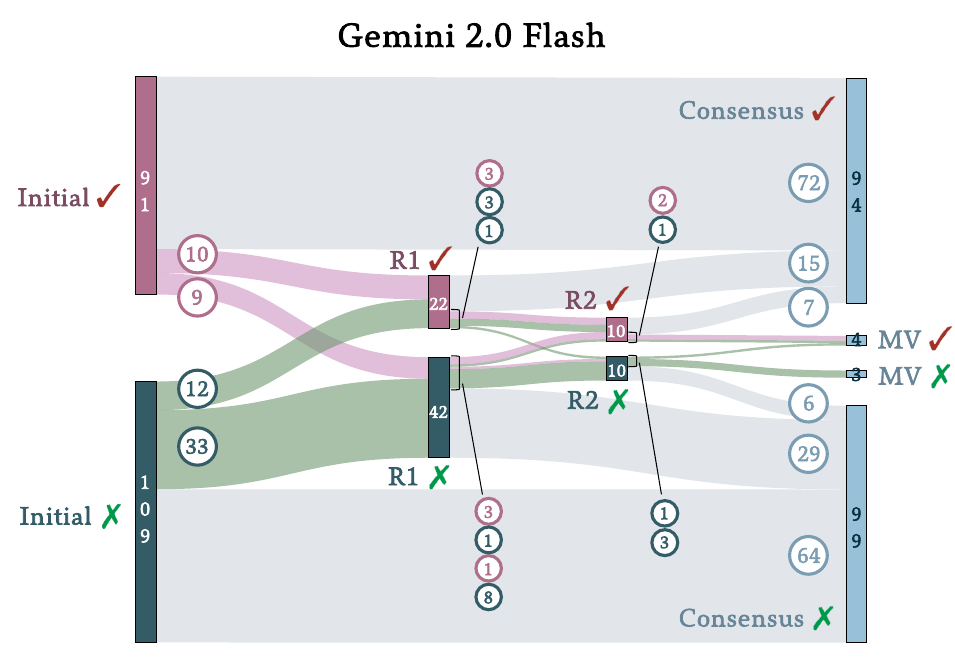}
    \end{minipage}
    \hfill
    \begin{minipage}{0.5\textwidth}
        \centering
        \includegraphics[width=\linewidth]{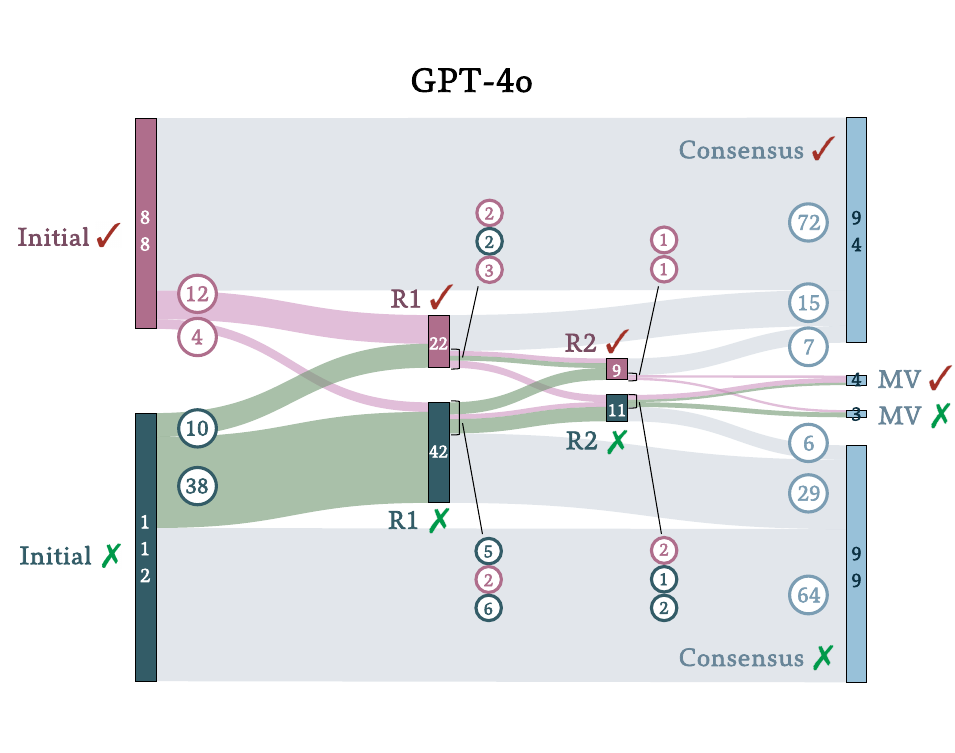}
    \end{minipage}
    
    \caption{Model behaviors within multi-agent discussion (1 reasoning model setting) on FoDS dataset. The 10 dark-colored nodes represent whether the current annotation is correct at different stages: \textcolor{darkr}{\rule{0.8em}{0.8em}} dark red (Initial \V, R1 \V, R2 \V), \textcolor{darkg}{\rule{0.8em}{0.8em}} dark green (Initial \X, R1 \X, R2 \X), and \textcolor{darkb}{\rule{0.8em}{0.8em}} dark blue (both Consensus \V/\X~and MV \V/\X). The 3 light-colored backgrounds indicate how models changed or retained their annotation during discussion: \textcolor{lightr}{\rule{0.8em}{0.8em}} light red (initial annotation is correct), \textcolor{lightg}{\rule{0.8em}{0.8em}} light green (initial annotation is wrong), and \textcolor{lightb}{\rule{0.8em}{0.8em}} light blue (all agents reach a consensus). The consensus (rightmost column) reflects the final agreed-upon annotation, which benefits from the collective refinement process. The numbers in the circles indicate the exact instance counts. Note that the number of instances shown in blue (both dark and light shades) is the same across all three models.}
    \label{fig:model-behavior}
\end{figure}

\section{Related Work}

The growing interest in LLMs as automated annotators has led to a surge of studies demonstrating their utility in diverse NLP annotation tasks~\citep{chiang2023can,vu2024foundational,ding2023gpt,choi2024gpts}.
Early investigations evaluate models like GPT-3.5 and ChatGPT on general-domain tasks reporting performance competitive with or even exceeding that of trained crowdworkers~\citep{gilardi2023chatgpt,alizadeh2023open,bansal2023large,zhu2023can}.
Methods such as active learning~\citep{zhang2023llmaaa} and reflection~\citep{he2023annollm} have shown promising results by leveraging the capabilities of LLMs for data annotation tasks.
However, these works primarily focus on general-domain tasks, leaving a gap in the exploration of domain-specific applications.

Another emerging trend involves using LLMs in collaborative or multi-agent settings~\citep{guo2024large,xi2025rise}.
Prior work has explored consensus-building through debate~\citep{du2023improving}, iterative discussion~\citep{chen2023reconcile}, and role-playing simulations~\citep{liang2023encouraging}, largely targeting reasoning tasks or factual correctness.
However, the application of such frameworks to emulate expert annotation group workflows remains underexplored.

Overall, we build upon previous research~\citep{tseng2024expert} by incorporating reasoning models, and extend the current literature by empirically evaluating both individual and multi-agent LLMs in expert annotation scenarios across finance, law, and biomedicine domains.
Our study bridges the gap between general-purpose annotation research and the more rigorous demands of domain-specific annotation tasks, offering a critical assessment of whether LLMs can serve as direct, out-of-the-box alternatives to expert annotators.

\section{Conclusion}
In this work, we investigate the research question: ``Can top-performing LLMs -- often perceived as having expert-level proficiency on academic and professional benchmarks -- serve as direct alternatives to human expert annotators?''
Our findings suggest that LLMs equipped with inference-time techniques (\textit{e.g.}, CoT, self-consistency) yield only marginal, and sometimes even negative, performance gains.
While reasoning models may perform slightly better, they do not show statistically significant improvements in most settings, and still fall short of human expert performance.
In addition, we introduce a multi-agent discussion framework that effectively enhances accuracy and consensus, outperforming individual LLMs.
Finally, we conduct a fine-grained analysis and identify specific model behaviors emerging during the multi-agent discussion process.

Our results -- spanning both single-LLM and multi-agent methods across three domains and five datasets -- indicates that performant LLMs may \textit{not} serve as a direct alternative for annotation tasks requiring domain expertise.
Relying solely on parametric knowledge to perform domain-specific, expert-level annotation remains a non-trivial challenge.
Given that these specialized domains often involve high-risk sectors, ensuring the precision and accuracy of annotated data is critical.
Overall, our findings highlight the gap between existing LLMs and human experts, underscoring the need for future efforts to develop reliable LLMs-as-Expert-Annotators systems.

\section*{Limitation and Future Works}

As we aim to provide direct insight and observation on whether top-performing LLMs can perform as expert annotators \textit{out-of-the-box}, we minimize efforts in prompt engineering.
Some works have demonstrated that, for specific scenarios, one can achieve sizable improvement through carefully-crafted prompts.
Consequently, our results may further benefit from a more exhaustive prompt optimization.

Another potential limitation is that we primarily focus on natural language understanding tasks with fixed label space.
Towards a more comprehensive evaluation, natural language generation tasks could be further incorporated.
Furthermore, all of our experimental settings involve zero-shot configurations using general-purpose chatbot LLMs.
To unveil more of the capabilities of LLMs in annotation tasks, future directions could explore few-shot settings, domain-specific or fine-tuned LLMs tailored to the annotation tasks, retrieval-augmented generation methods, or a promising human-LLM hybrid annotation schema.

\section*{Acknowledgments}
This work was supported by National Science and Technology Council, Taiwan, under grants NSTC 113-2634-F-002-003- and 114-2221-E-002-070-MY3, and Ministry of Education (MOE), Taiwan, under grant NTU-114L900901. 
The work of Chung-Chi Chen was supported in part by AIST policy-based budget project ``R\&D on Generative AI Foundation Models for the Physical Domain.''

\bibliography{main}

\begin{thebibliography}{48}
\providecommand{\natexlab}[1]{#1}
\providecommand{\url}[1]{\texttt{#1}}
\expandafter\ifx\csname urlstyle\endcsname\relax
  \providecommand{\doi}[1]{doi: #1}\else
  \providecommand{\doi}{doi: \begingroup \urlstyle{rm}\Url}\fi

\bibitem[Achiam et~al.(2023)Achiam, Adler, Agarwal, Ahmad, Akkaya, Aleman, Almeida, Altenschmidt, Altman, Anadkat, et~al.]{achiam2023gpt}
Josh Achiam, Steven Adler, Sandhini Agarwal, Lama Ahmad, Ilge Akkaya, Florencia~Leoni Aleman, Diogo Almeida, Janko Altenschmidt, Sam Altman, Shyamal Anadkat, et~al.
\newblock Gpt-4 technical report.
\newblock \emph{arXiv preprint arXiv:2303.08774}, 2023.

\bibitem[Alizadeh et~al.(2023)Alizadeh, Kubli, Samei, Dehghani, Bermeo, Korobeynikova, and Gilardi]{alizadeh2023open}
Meysam Alizadeh, Ma{\"e}l Kubli, Zeynab Samei, Shirin Dehghani, Juan~Diego Bermeo, Maria Korobeynikova, and Fabrizio Gilardi.
\newblock Open-source large language models outperform crowd workers and approach chatgpt in text-annotation tasks.
\newblock \emph{arXiv preprint arXiv:2307.02179}, 2023.

\bibitem[Anthropic(2024)]{anthropic2024claude}
AI~Anthropic.
\newblock The claude 3 model family: Opus, sonnet, haiku.
\newblock \emph{Claude-3 Model Card}, 2024.

\bibitem[Anthropic(2025)]{anthropic2025claude}
AI~Anthropic.
\newblock Claude 3.7 sonnet and claude code.
\newblock 2025.
\newblock URL \url{https://www.anthropic.com/news/claude-3-7-sonnet}.

\bibitem[Bansal \& Sharma(2023)Bansal and Sharma]{bansal2023large}
Parikshit Bansal and Amit Sharma.
\newblock Large language models as annotators: Enhancing generalization of nlp models at minimal cost.
\newblock \emph{arXiv preprint arXiv:2306.15766}, 2023.

\bibitem[Bozdag et~al.(2025)Bozdag, Mehri, Yang, Ha, Cheng, Durmus, You, Ji, Tur, and Hakkani-T{\"u}r]{bozdag2025must}
Nimet~Beyza Bozdag, Shuhaib Mehri, Xiaocheng Yang, Hyeonjeong Ha, Zirui Cheng, Esin Durmus, Jiaxuan You, Heng Ji, Gokhan Tur, and Dilek Hakkani-T{\"u}r.
\newblock Must read: A systematic survey of computational persuasion.
\newblock \emph{arXiv preprint arXiv:2505.07775}, 2025.

\bibitem[Callanan et~al.(2023)Callanan, Mbakwe, Papadimitriou, Pei, Sibue, Zhu, Ma, Liu, and Shah]{callanan2023can}
Ethan Callanan, Amarachi Mbakwe, Antony Papadimitriou, Yulong Pei, Mathieu Sibue, Xiaodan Zhu, Zhiqiang Ma, Xiaomo Liu, and Sameena Shah.
\newblock Can gpt models be financial analysts? an evaluation of chatgpt and gpt-4 on mock cfa exams.
\newblock \emph{arXiv preprint arXiv:2310.08678}, 2023.

\bibitem[Chen et~al.(2023)Chen, Saha, and Bansal]{chen2023reconcile}
Justin Chih-Yao Chen, Swarnadeep Saha, and Mohit Bansal.
\newblock Reconcile: Round-table conference improves reasoning via consensus among diverse llms.
\newblock \emph{arXiv preprint arXiv:2309.13007}, 2023.

\bibitem[Chen et~al.(2021)Chen, Tworek, Jun, Yuan, Pinto, Kaplan, Edwards, Burda, Joseph, Brockman, et~al.]{chen2021evaluating}
Mark Chen, Jerry Tworek, Heewoo Jun, Qiming Yuan, Henrique Ponde de~Oliveira Pinto, Jared Kaplan, Harri Edwards, Yuri Burda, Nicholas Joseph, Greg Brockman, et~al.
\newblock Evaluating large language models trained on code.
\newblock \emph{arXiv preprint arXiv:2107.03374}, 2021.

\bibitem[Chiang \& Lee(2023)Chiang and Lee]{chiang2023can}
Cheng-Han Chiang and Hung-Yi Lee.
\newblock Can large language models be an alternative to human evaluations?
\newblock In \emph{Proceedings of the 61st Annual Meeting of the Association for Computational Linguistics (Volume 1: Long Papers)}, pp.\  15607--15631, 2023.

\bibitem[Choi et~al.(2021)Choi, Hickman, Monahan, and Schwarcz]{choi2021chatgpt}
Jonathan~H Choi, Kristin~E Hickman, Amy~B Monahan, and Daniel Schwarcz.
\newblock Chatgpt goes to law school.
\newblock \emph{J. Legal Educ.}, 71:\penalty0 387, 2021.

\bibitem[Choi et~al.(2024)Choi, Lee, Jin, and Kim]{choi2024gpts}
Juhwan Choi, Eunju Lee, Kyohoon Jin, and YoungBin Kim.
\newblock Gpts are multilingual annotators for sequence generation tasks.
\newblock \emph{arXiv preprint arXiv:2402.05512}, 2024.

\bibitem[Ding et~al.(2023)Ding, Qin, Liu, Chia, Li, Joty, and Bing]{ding2023gpt}
Bosheng Ding, Chengwei Qin, Linlin Liu, Yew~Ken Chia, Boyang Li, Shafiq Joty, and Lidong Bing.
\newblock Is gpt-3 a good data annotator?
\newblock In \emph{Proceedings of the 61st Annual Meeting of the Association for Computational Linguistics (Volume 1: Long Papers)}, pp.\  11173--11195, 2023.

\bibitem[Du et~al.(2023)Du, Li, Torralba, Tenenbaum, and Mordatch]{du2023improving}
Yilun Du, Shuang Li, Antonio Torralba, Joshua~B Tenenbaum, and Igor Mordatch.
\newblock Improving factuality and reasoning in language models through multiagent debate.
\newblock \emph{arXiv preprint arXiv:2305.14325}, 2023.

\bibitem[Durmus et~al.(2024)Durmus, Lovitt, Tamkin, Ritchie, Clark, and Ganguli]{durmus2024persuasion}
Esin Durmus, Liane Lovitt, Alex Tamkin, Stuart Ritchie, Jack Clark, and Deep Ganguli.
\newblock Measuring the persuasiveness of language models, 2024.
\newblock URL \url{https://www.anthropic.com/news/measuring-model-persuasiveness}.

\bibitem[Fleiss(1971)]{fleiss1971measuring}
Joseph~L Fleiss.
\newblock Measuring nominal scale agreement among many raters.
\newblock \emph{Psychological bulletin}, 76\penalty0 (5):\penalty0 378, 1971.

\bibitem[Gilardi et~al.(2023)Gilardi, Alizadeh, and Kubli]{gilardi2023chatgpt}
Fabrizio Gilardi, Meysam Alizadeh, and Ma{\"e}l Kubli.
\newblock Chatgpt outperforms crowd workers for text-annotation tasks.
\newblock \emph{Proceedings of the National Academy of Sciences}, 120\penalty0 (30):\penalty0 e2305016120, 2023.

\bibitem[Google(2024)]{google2024gemini2.0}
Google.
\newblock Introducing gemini 2.0: our new ai model for the agentic era.
\newblock 2024.
\newblock URL \url{https://blog.google/technology/google-deepmind/google-gemini-ai-update-december-2024/#ceo-message}.

\bibitem[Guha et~al.(2024)Guha, Nyarko, Ho, R{\'e}, Chilton, Chohlas-Wood, Peters, Waldon, Rockmore, Zambrano, et~al.]{guha2024legalbench}
Neel Guha, Julian Nyarko, Daniel Ho, Christopher R{\'e}, Adam Chilton, Alex Chohlas-Wood, Austin Peters, Brandon Waldon, Daniel Rockmore, Diego Zambrano, et~al.
\newblock Legalbench: A collaboratively built benchmark for measuring legal reasoning in large language models.
\newblock \emph{Advances in Neural Information Processing Systems}, 36, 2024.

\bibitem[Guo et~al.(2024)Guo, Chen, Wang, Chang, Pei, Chawla, Wiest, and Zhang]{guo2024large}
Taicheng Guo, Xiuying Chen, Yaqi Wang, Ruidi Chang, Shichao Pei, Nitesh~V Chawla, Olaf Wiest, and Xiangliang Zhang.
\newblock Large language model based multi-agents: A survey of progress and challenges.
\newblock \emph{arXiv preprint arXiv:2402.01680}, 2024.

\bibitem[He et~al.(2023)He, Lin, Gong, Jin, Zhang, Lin, Jiao, Yiu, Duan, Chen, et~al.]{he2023annollm}
Xingwei He, Zhenghao Lin, Yeyun Gong, Alex Jin, Hang Zhang, Chen Lin, Jian Jiao, Siu~Ming Yiu, Nan Duan, Weizhu Chen, et~al.
\newblock Annollm: Making large language models to be better crowdsourced annotators.
\newblock \emph{arXiv preprint arXiv:2303.16854}, 2023.

\bibitem[Hendrycks et~al.(2020)Hendrycks, Burns, Basart, Zou, Mazeika, Song, and Steinhardt]{hendrycks2020measuring}
Dan Hendrycks, Collin Burns, Steven Basart, Andy Zou, Mantas Mazeika, Dawn Song, and Jacob Steinhardt.
\newblock Measuring massive multitask language understanding.
\newblock In \emph{International Conference on Learning Representations}, 2020.

\bibitem[Hendrycks et~al.(2021)Hendrycks, Burns, Chen, and Ball]{hendrycks2021cuad}
Dan Hendrycks, Collin Burns, Anya Chen, and Spencer Ball.
\newblock Cuad: An expert-annotated nlp dataset for legal contract review.
\newblock \emph{arXiv preprint arXiv:2103.06268}, 2021.

\bibitem[Huang et~al.(2020)Huang, Huang, Ding, Hsu, and Giles]{huang2020coda}
Ting-Hao~Kenneth Huang, Chieh-Yang Huang, Chien-Kuang~Cornelia Ding, Yen-Chia Hsu, and C~Lee Giles.
\newblock Coda-19: Using a non-expert crowd to annotate research aspects on 10,000+ abstracts in the covid-19 open research dataset.
\newblock In \emph{ACL 2020 Workshop on Natural Language Processing for COVID-19 (NLP-COVID)}, 2020.

\bibitem[Jin et~al.(2019)Jin, Dhingra, Liu, Cohen, and Lu]{jin2019pubmedqa}
Qiao Jin, Bhuwan Dhingra, Zhengping Liu, William Cohen, and Xinghua Lu.
\newblock Pubmedqa: A dataset for biomedical research question answering.
\newblock In \emph{Proceedings of the 2019 Conference on Empirical Methods in Natural Language Processing and the 9th International Joint Conference on Natural Language Processing (EMNLP-IJCNLP)}, pp.\  2567--2577, 2019.

\bibitem[Katz et~al.(2024)Katz, Bommarito, Gao, and Arredondo]{katz2024gpt}
Daniel~Martin Katz, Michael~James Bommarito, Shang Gao, and Pablo Arredondo.
\newblock Gpt-4 passes the bar exam.
\newblock \emph{Philosophical Transactions of the Royal Society A}, 382\penalty0 (2270):\penalty0 20230254, 2024.

\bibitem[Kaur et~al.(2023)Kaur, Smiley, Gupta, Sain, Wang, Siddagangappa, Aguda, and Shah]{kaur2023refind}
Simerjot Kaur, Charese Smiley, Akshat Gupta, Joy Sain, Dongsheng Wang, Suchetha Siddagangappa, Toyin Aguda, and Sameena Shah.
\newblock Refind: Relation extraction financial dataset.
\newblock In \emph{Proceedings of the 46th International ACM SIGIR Conference on Research and Development in Information Retrieval}, pp.\  3054--3063, 2023.

\bibitem[Kojima et~al.(2022)Kojima, Gu, Reid, Matsuo, and Iwasawa]{kojima2022large}
Takeshi Kojima, Shixiang~Shane Gu, Machel Reid, Yutaka Matsuo, and Yusuke Iwasawa.
\newblock Large language models are zero-shot reasoners.
\newblock \emph{Advances in neural information processing systems}, 35:\penalty0 22199--22213, 2022.

\bibitem[Liang et~al.(2023)Liang, He, Jiao, Wang, Wang, Wang, Yang, Tu, and Shi]{liang2023encouraging}
Tian Liang, Zhiwei He, Wenxiang Jiao, Xing Wang, Yan Wang, Rui Wang, Yujiu Yang, Zhaopeng Tu, and Shuming Shi.
\newblock Encouraging divergent thinking in large language models through multi-agent debate.
\newblock \emph{arXiv preprint arXiv:2305.19118}, 2023.

\bibitem[Madaan et~al.(2024)Madaan, Tandon, Gupta, Hallinan, Gao, Wiegreffe, Alon, Dziri, Prabhumoye, Yang, et~al.]{madaan2024self}
Aman Madaan, Niket Tandon, Prakhar Gupta, Skyler Hallinan, Luyu Gao, Sarah Wiegreffe, Uri Alon, Nouha Dziri, Shrimai Prabhumoye, Yiming Yang, et~al.
\newblock Self-refine: Iterative refinement with self-feedback.
\newblock \emph{Advances in Neural Information Processing Systems}, 36, 2024.

\bibitem[McNemar(1947)]{mcnemar1947note}
Quinn McNemar.
\newblock Note on the sampling error of the difference between correlated proportions or percentages.
\newblock \emph{Psychometrika}, 12\penalty0 (2):\penalty0 153--157, 1947.

\bibitem[OpenAI(2024)]{openai2024gpt4o}
OpenAI.
\newblock Hello gpt4-o.
\newblock 2024.
\newblock URL \url{https://openai.com/index/hello-gpt-4o/}.

\bibitem[OpenAI(2025)]{openai2025o3mini}
OpenAI.
\newblock Openai o3-mini.
\newblock 2025.
\newblock URL \url{https://openai.com/index/openai-o3-mini/}.

\bibitem[Reid et~al.(2024)Reid, Savinov, Teplyashin, Lepikhin, Lillicrap, Alayrac, Soricut, Lazaridou, Firat, Schrittwieser, et~al.]{reid2024gemini}
Machel Reid, Nikolay Savinov, Denis Teplyashin, Dmitry Lepikhin, Timothy Lillicrap, Jean-baptiste Alayrac, Radu Soricut, Angeliki Lazaridou, Orhan Firat, Julian Schrittwieser, et~al.
\newblock Gemini 1.5: Unlocking multimodal understanding across millions of tokens of context.
\newblock \emph{arXiv preprint arXiv:2403.05530}, 2024.

\bibitem[Rein et~al.(2023)Rein, Hou, Stickland, Petty, Pang, Dirani, Michael, and Bowman]{rein2023gpqa}
David Rein, Betty~Li Hou, Asa~Cooper Stickland, Jackson Petty, Richard~Yuanzhe Pang, Julien Dirani, Julian Michael, and Samuel~R Bowman.
\newblock Gpqa: A graduate-level google-proof q\&a benchmark.
\newblock \emph{arXiv preprint arXiv:2311.12022}, 2023.

\bibitem[Shah et~al.(2023)Shah, Paturi, and Chava]{shah2023trillion}
Agam Shah, Suvan Paturi, and Sudheer Chava.
\newblock Trillion dollar words: A new financial dataset, task \& market analysis.
\newblock In \emph{Proceedings of the 61st Annual Meeting of the Association for Computational Linguistics (Volume 1: Long Papers)}, pp.\  6664--6679, 2023.

\bibitem[Singhal et~al.(2023{\natexlab{a}})Singhal, Azizi, Tu, Mahdavi, Wei, Chung, Scales, Tanwani, Cole-Lewis, Pfohl, et~al.]{singhal2023large}
Karan Singhal, Shekoofeh Azizi, Tao Tu, S~Sara Mahdavi, Jason Wei, Hyung~Won Chung, Nathan Scales, Ajay Tanwani, Heather Cole-Lewis, Stephen Pfohl, et~al.
\newblock Large language models encode clinical knowledge.
\newblock \emph{Nature}, 620\penalty0 (7972):\penalty0 172--180, 2023{\natexlab{a}}.

\bibitem[Singhal et~al.(2023{\natexlab{b}})Singhal, Tu, Gottweis, Sayres, Wulczyn, Hou, Clark, Pfohl, Cole-Lewis, Neal, et~al.]{singhal2023towards}
Karan Singhal, Tao Tu, Juraj Gottweis, Rory Sayres, Ellery Wulczyn, Le~Hou, Kevin Clark, Stephen Pfohl, Heather Cole-Lewis, Darlene Neal, et~al.
\newblock Towards expert-level medical question answering with large language models.
\newblock \emph{arXiv preprint arXiv:2305.09617}, 2023{\natexlab{b}}.

\bibitem[Tan et~al.(2024)Tan, Beigi, Wang, Guo, Bhattacharjee, Jiang, Karami, Li, Cheng, and Liu]{tan2024large}
Zhen Tan, Alimohammad Beigi, Song Wang, Ruocheng Guo, Amrita Bhattacharjee, Bohan Jiang, Mansooreh Karami, Jundong Li, Lu~Cheng, and Huan Liu.
\newblock Large language models for data annotation: A survey.
\newblock \emph{arXiv preprint arXiv:2402.13446}, 2024.

\bibitem[Tseng et~al.(2024{\natexlab{a}})Tseng, Chen, Chen, and Chen]{tseng2024expert}
Yu-Min Tseng, Wei-Lin Chen, Chung-Chi Chen, and Hsin-Hsi Chen.
\newblock Are expert-level language models expert-level annotators?
\newblock \emph{arXiv preprint arXiv:2410.03254}, 2024{\natexlab{a}}.

\bibitem[Tseng et~al.(2024{\natexlab{b}})Tseng, Huang, Hsiao, Hsu, Foo, Huang, and Chen]{tseng2024two}
Yu-Min Tseng, Yu-Chao Huang, Teng-Yun Hsiao, Yu-Ching Hsu, Jia-Yin Foo, Chao-Wei Huang, and Yun-Nung Chen.
\newblock Two tales of persona in llms: A survey of role-playing and personalization.
\newblock \emph{arXiv preprint arXiv:2406.01171}, 2024{\natexlab{b}}.

\bibitem[Vu et~al.(2024)Vu, Krishna, Alzubi, Tar, Faruqui, and Sung]{vu2024foundational}
Tu~Vu, Kalpesh Krishna, Salaheddin Alzubi, Chris Tar, Manaal Faruqui, and Yun-Hsuan Sung.
\newblock Foundational autoraters: Taming large language models for better automatic evaluation.
\newblock In \emph{Proceedings of the 2024 Conference on Empirical Methods in Natural Language Processing}, pp.\  17086--17105, 2024.

\bibitem[Wang et~al.(2020)Wang, Lo, Chandrasekhar, Reas, Yang, Burdick, Eide, Funk, Katsis, Kinney, et~al.]{wang2020cord}
Lucy~Lu Wang, Kyle Lo, Yoganand Chandrasekhar, Russell Reas, Jiangjiang Yang, Douglas Burdick, Darrin Eide, Kathryn Funk, Yannis Katsis, Rodney Kinney, et~al.
\newblock Cord-19: The covid-19 open research dataset.
\newblock \emph{ArXiv}, 2020.

\bibitem[Wang et~al.(2022)Wang, Wei, Schuurmans, Le, Chi, Narang, Chowdhery, and Zhou]{wang2022self}
Xuezhi Wang, Jason Wei, Dale Schuurmans, Quoc~V Le, Ed~H Chi, Sharan Narang, Aakanksha Chowdhery, and Denny Zhou.
\newblock Self-consistency improves chain of thought reasoning in language models.
\newblock In \emph{The Eleventh International Conference on Learning Representations}, 2022.

\bibitem[Wei et~al.(2022)Wei, Wang, Schuurmans, Bosma, Xia, Chi, Le, Zhou, et~al.]{wei2022chain}
Jason Wei, Xuezhi Wang, Dale Schuurmans, Maarten Bosma, Fei Xia, Ed~Chi, Quoc~V Le, Denny Zhou, et~al.
\newblock Chain-of-thought prompting elicits reasoning in large language models.
\newblock \emph{Advances in neural information processing systems}, 35:\penalty0 24824--24837, 2022.

\bibitem[Xi et~al.(2025)Xi, Chen, Guo, He, Ding, Hong, Zhang, Wang, Jin, Zhou, et~al.]{xi2025rise}
Zhiheng Xi, Wenxiang Chen, Xin Guo, Wei He, Yiwen Ding, Boyang Hong, Ming Zhang, Junzhe Wang, Senjie Jin, Enyu Zhou, et~al.
\newblock The rise and potential of large language model based agents: A survey.
\newblock \emph{Science China Information Sciences}, 68\penalty0 (2):\penalty0 121101, 2025.

\bibitem[Zhang et~al.(2023)Zhang, Li, Ma, Zhou, and Zou]{zhang2023llmaaa}
Ruoyu Zhang, Yanzeng Li, Yongliang Ma, Ming Zhou, and Lei Zou.
\newblock Llmaaa: Making large language models as active annotators.
\newblock \emph{arXiv preprint arXiv:2310.19596}, 2023.

\bibitem[Zhu et~al.(2023)Zhu, Zhang, Haq, Hui, and Tyson]{zhu2023can}
Yiming Zhu, Peixian Zhang, Ehsan-Ul Haq, Pan Hui, and Gareth Tyson.
\newblock Can chatgpt reproduce human-generated labels? a study of social computing tasks.
\newblock \emph{arXiv preprint arXiv:2304.10145}, 2023.

\end{thebibliography}
\bibliographystyle{colm2025_conference}

\appendix

\section{Datasets}\label{appendix-dataset}

\subsection{Finance}
\textbf{REFinD}~\citep{kaur2023refind} is the largest relation extraction dataset over financial documents, comprising 8 entity pairs and 22 relations, with labels reviewed by financial experts.
In this task, annotators are tasked to select the relation type between finance-specific entity pairs, such as [\textit{person}] is an employee of [\textit{organization}].

\textbf{FOMC}~\citep{shah2023trillion} is constructed for identifying sentiments about the future monetary policy stances, annotated by experts with a correlated financial knowledge.
The labels of this annotation task are Dovish, Hawkish, and Neutral, where a Dovish sentence indicates easing and a Hawkish sentence indicates tightening.

\subsection{Law}
\textbf{CUAD} (Contract Understanding Atticus Dataset;~\citealp{hendrycks2021cuad}) consists of legal contracts with extensive annotations from legal experts, created with a year-long effort by dozens of law student annotators, lawyers, and machine learning researchers.
The annotation task is to label 41 types out of legal clauses, classified into 5 answer categories, that are considered important in contract review related to corporate transactions.
We manually use ``Yes/No'' answer category to construct our annotation task as the identification of 32 types of clauses.

\textbf{FoDS} (Function of Decision Section dataset;~\citealp{guha2024legalbench}) comprises one-paragraph excerpts from legal decisions, annotated by legal professionals.
In this task, annotators are tasked to review a legal decision and identify one out of seven function categories that each section (\textit{i.e.}, excerpt) of the decision serves.

\subsection{Biomedicine}
\textbf{CODA-19} (COVID-19 Research Aspect Dataset;~\citealp{huang2020coda}) codes each segment aspect of English abstracts in the COVID-19 Open Research Dataset~\citep{wang2020cord}.
In this task, annotators are tasked to label each segment as \textit{background}, \textit{purpose}, \textit{method}, \textit{finding/contribution}, or \textit{other} sections.
To ensure the quality of the labels, we only adopt instances annotated by biomedical experts.
We provide data statistics of the five existing specialized datasets in Table~\ref{tab:dataset-statistic}.

\section{Annotation Guidelines}\label{subsec:appendix-guideline}
\begin{figure*}[t]
  \centering
  \includegraphics[width=\linewidth]{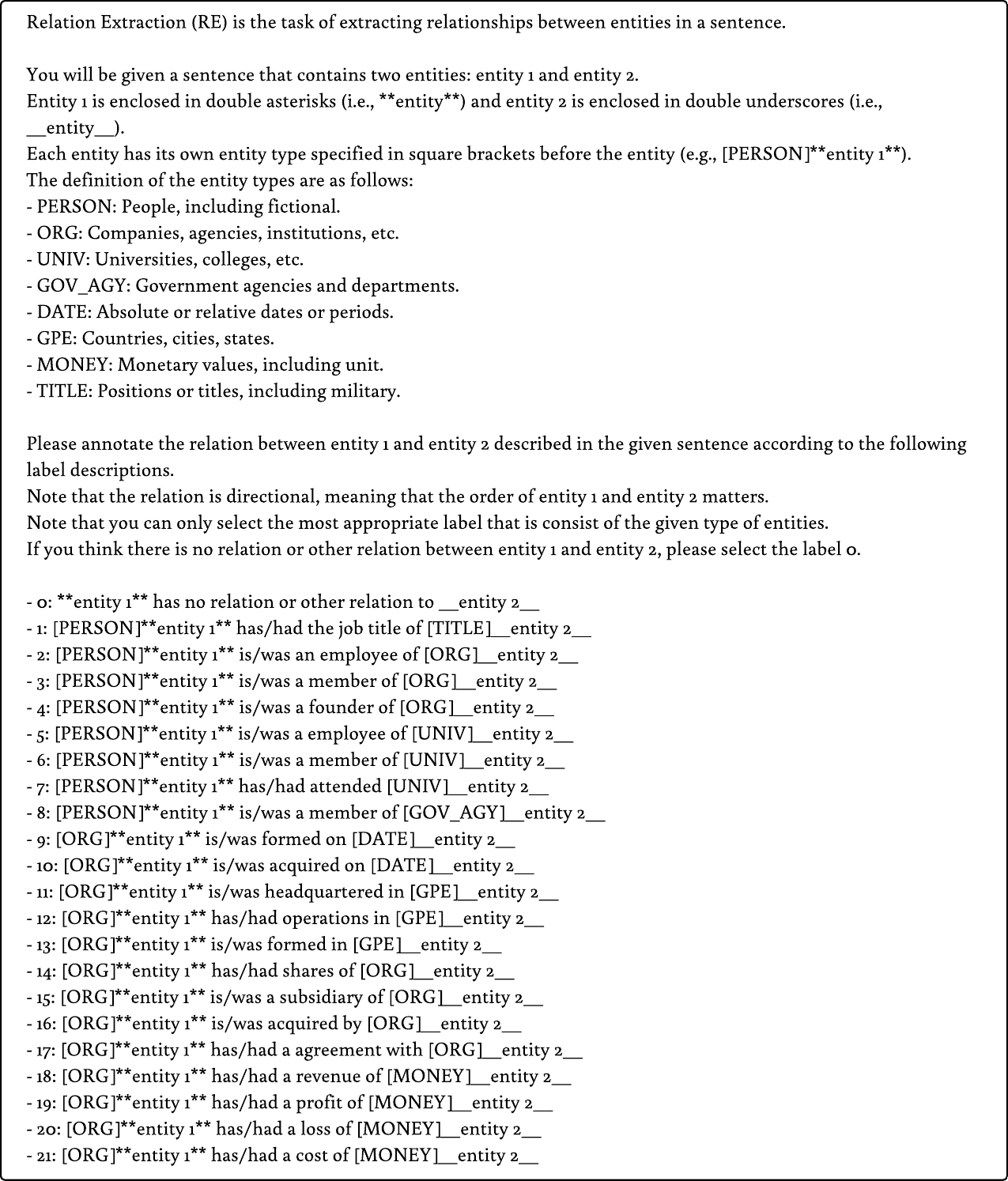}
  \caption{The annotation guideline of REFinD dataset.}
  \label{fig:guideline-refind}
\end{figure*}

\begin{figure*}[t]
  \centering
  \includegraphics[width=\linewidth]{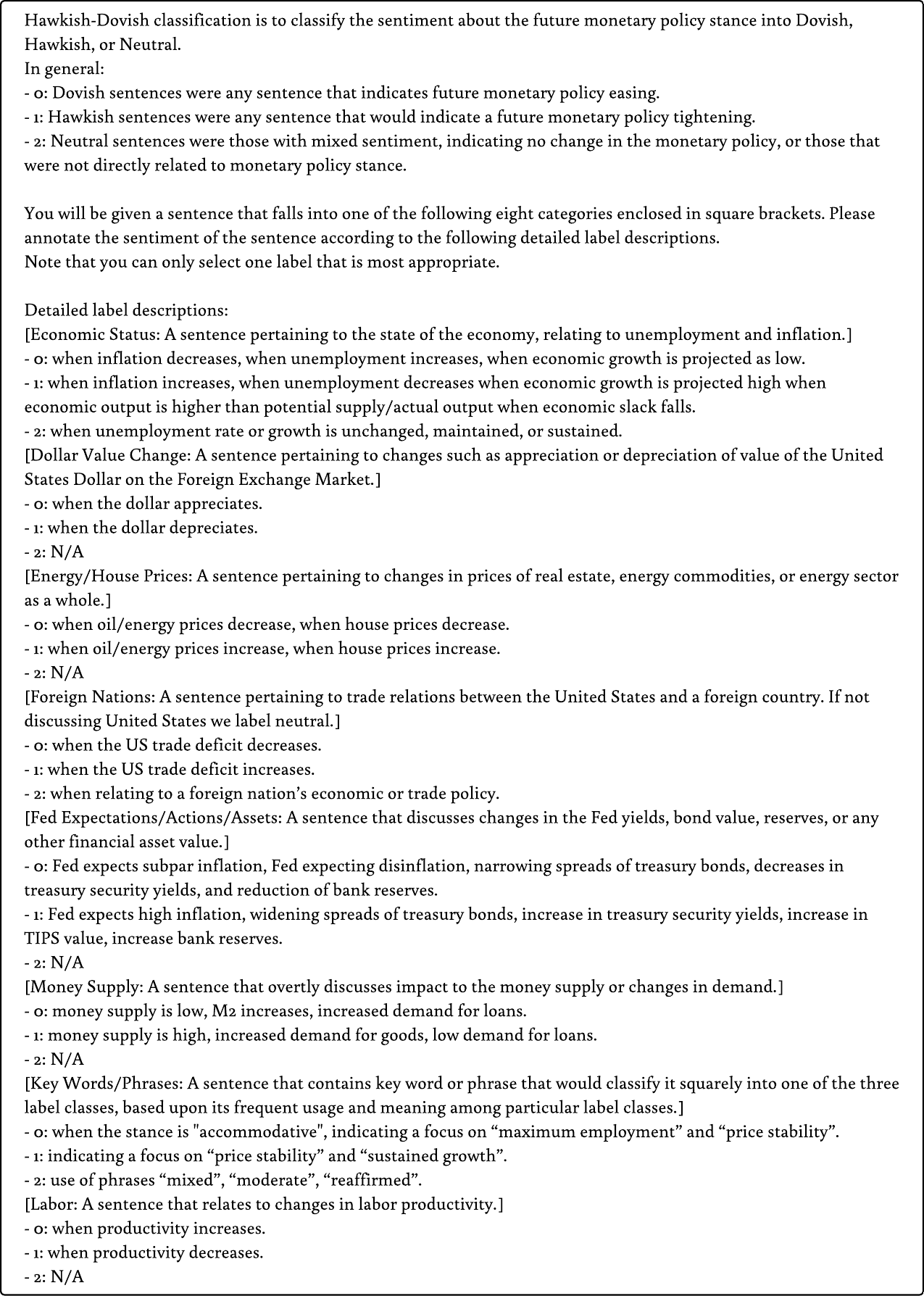}
  \caption{The annotation guideline of FOMC dataset.}
  \label{fig:guideline-fomc}
\end{figure*}

\begin{figure*}[t]
  \centering
  \includegraphics[width=\linewidth]{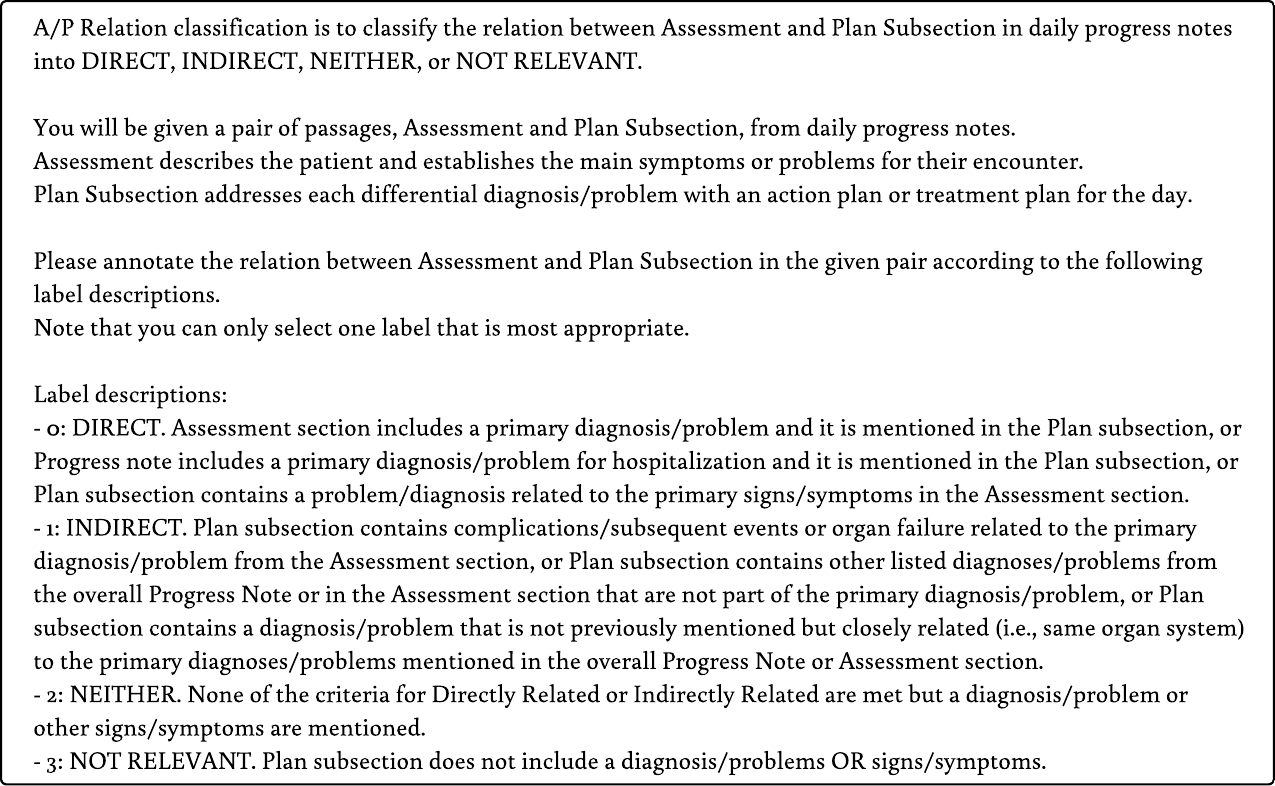}
  \caption{The annotation guideline of AP-Relation dataset.}
  \label{fig:guideline-ap-relation}
\end{figure*}

\begin{figure*}[t]
  \centering
  \includegraphics[width=\linewidth]{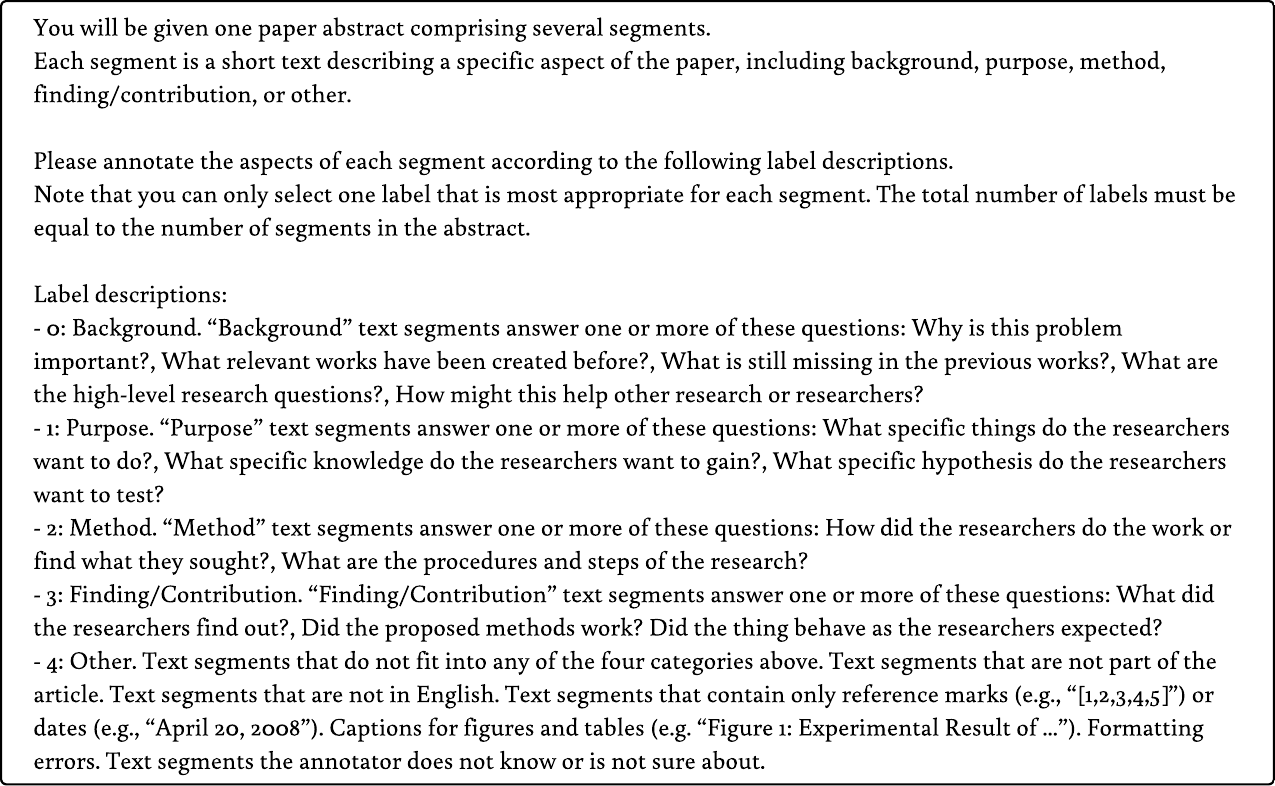}
  \caption{The annotation guideline of CODA-19 dataset.}
  \label{fig:guideline-coda-19}
\end{figure*}

\begin{figure*}[t]
  \centering
  \includegraphics[width=\linewidth]{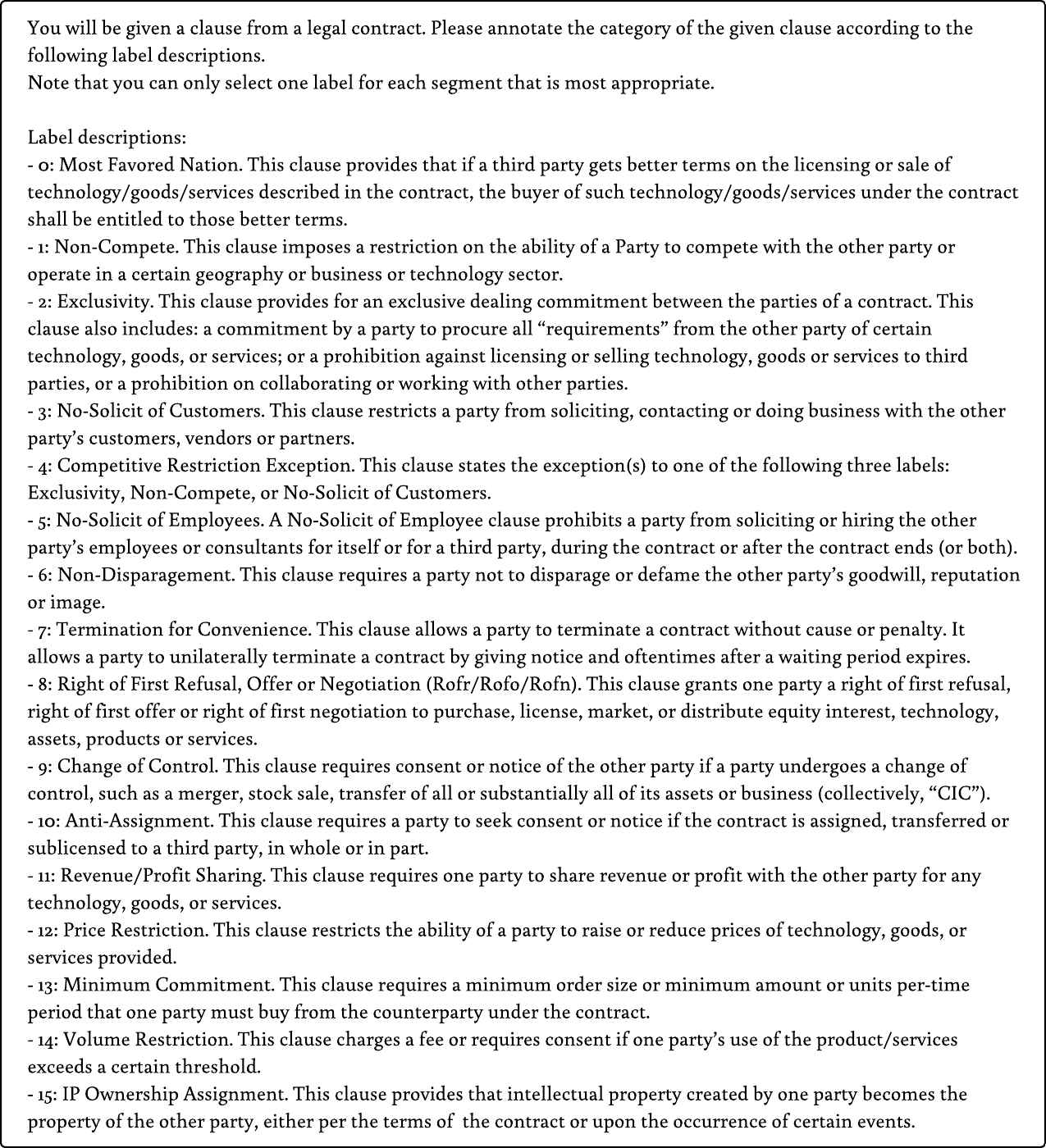}
  \caption{The annotation guideline of CUAD dataset (1-1).}
  \label{fig:guideline-coda-19-1}
\end{figure*}

\begin{figure*}[t]
  \centering
  \includegraphics[width=\linewidth]{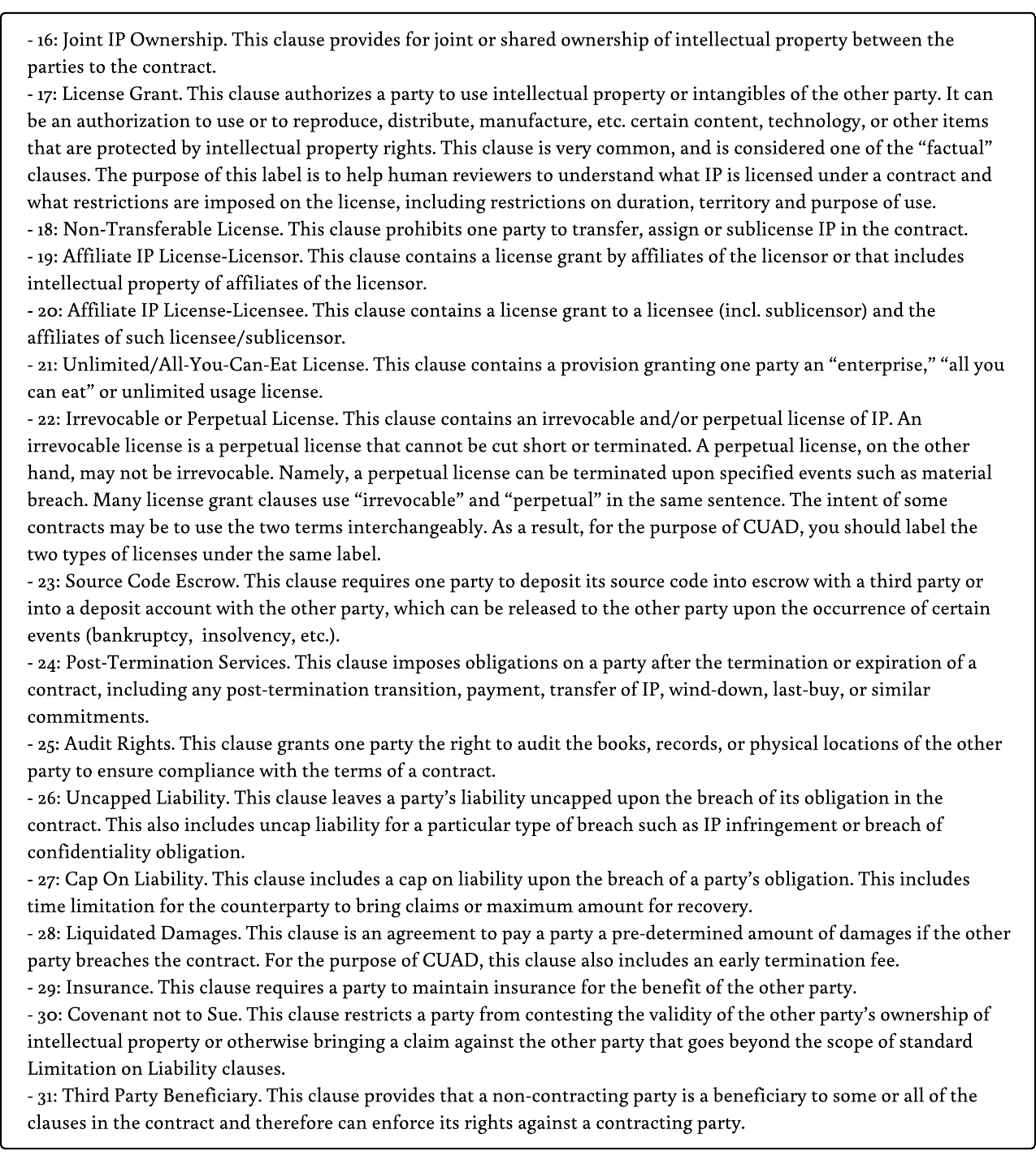}
  \caption{The annotation guideline of CUAD dataset (1-2).}
  \label{fig:guideline-coda-19-2}
\end{figure*}

\begin{figure*}[t]
  \centering
  \includegraphics[width=\linewidth]{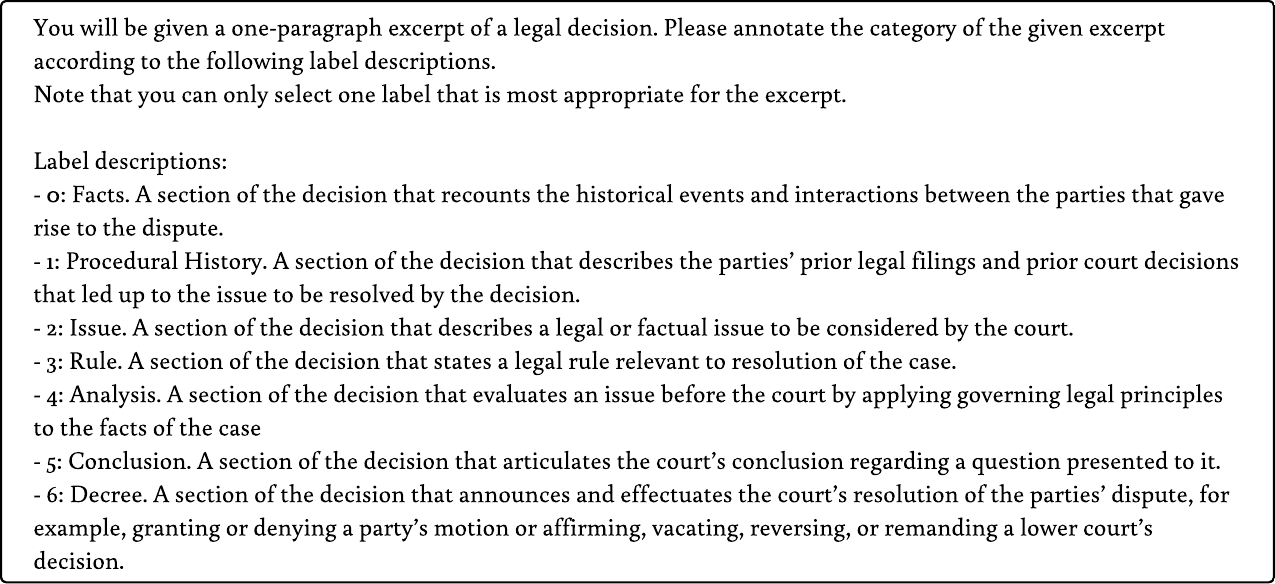}
  \caption{The annotation guideline of FoDS dataset.}
  \label{fig:guideline-fods}
\end{figure*}

We provide annotation guidelines of each dataset from Figure~\ref{fig:guideline-refind} to Figure~\ref{fig:guideline-fods}.

\section{Prompt Templates}\label{subsec:appendix-prompt}
\begin{figure*}[b]
  \centering
  \includegraphics[width=\linewidth]{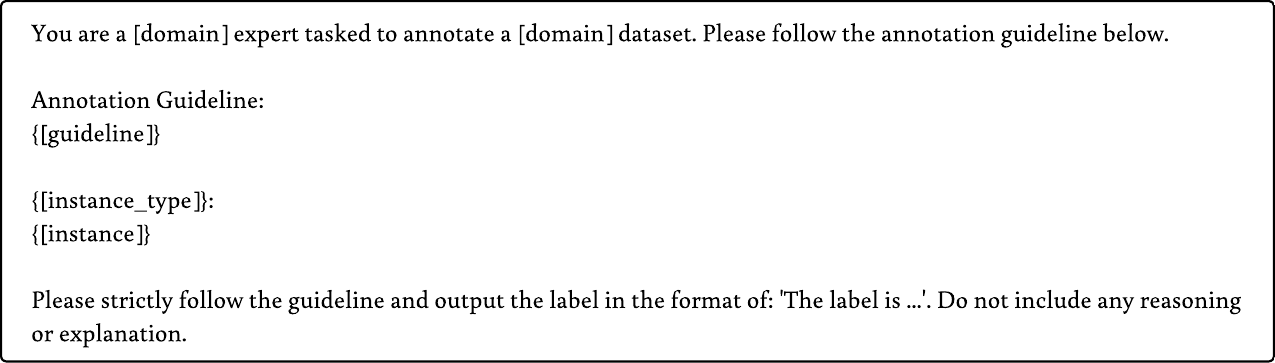}
  \caption{Vanilla prompt template.}
  \label{fig:template-vanilla}
\end{figure*}

\begin{figure*}[b]
  \centering
  \includegraphics[width=\linewidth]{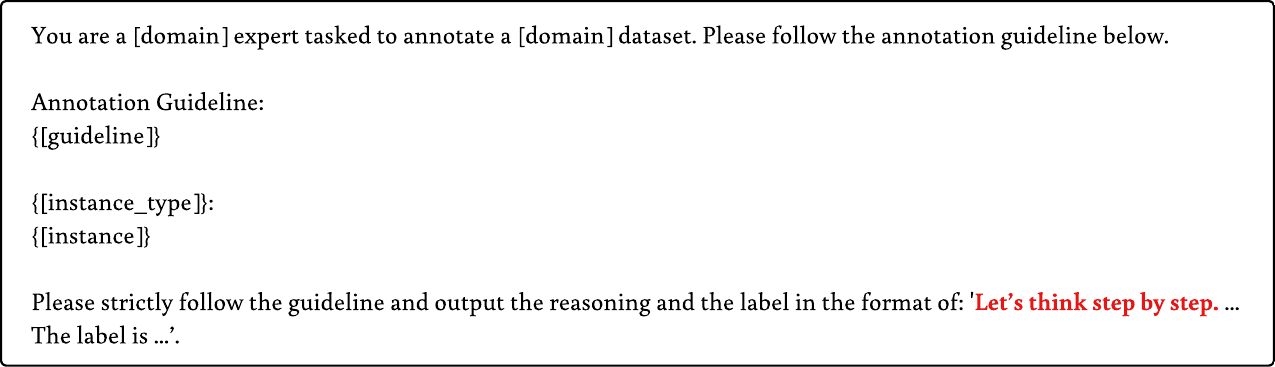}
  \caption{Chain-of-Thought prompt template.}
  \label{fig:template-cot}
\end{figure*}

\begin{figure*}[t]
  \centering
  \includegraphics[width=\linewidth]{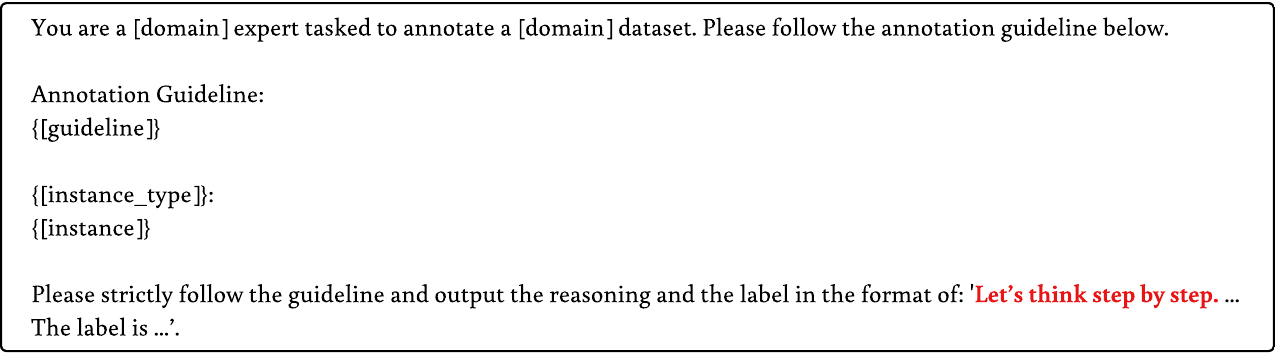}
  \caption{Self-Refine prompt template. Step 1: Generate.}
  \label{fig:template-self-refine-1}
\end{figure*}

\begin{figure*}[t]
  \centering
  \includegraphics[width=\linewidth]{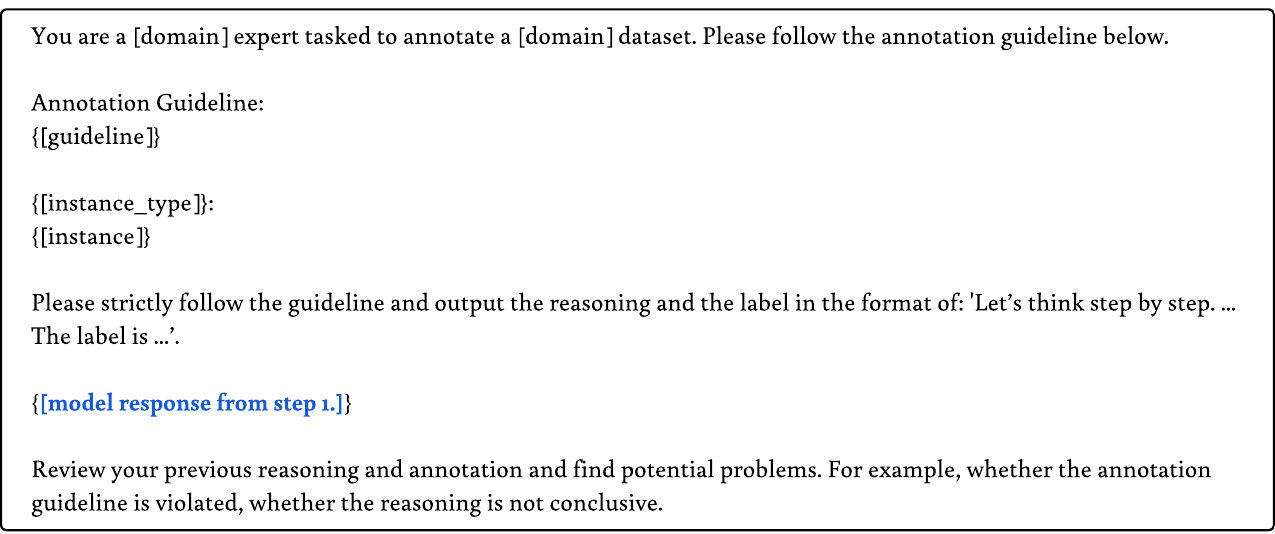}
  \caption{Self-Refine prompt template. Step 2: Review.}
  \label{fig:template-self-refine-2}
\end{figure*}

\begin{figure*}[t]
  \centering
  \includegraphics[width=\linewidth]{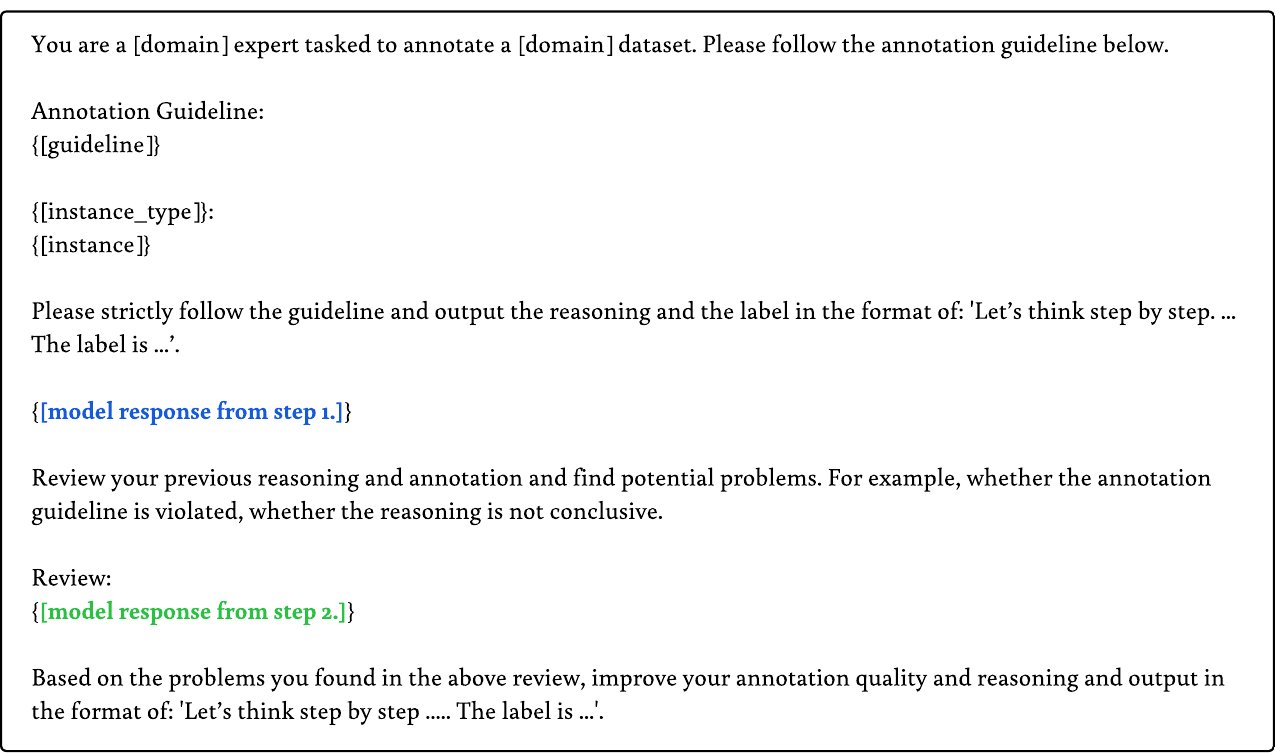}
  \caption{Self-Refine prompt template. Step 3: Refine.}
  \label{fig:template-self-refine-3}
\end{figure*}

\begin{figure*}[t]
  \centering
  \includegraphics[width=\linewidth]{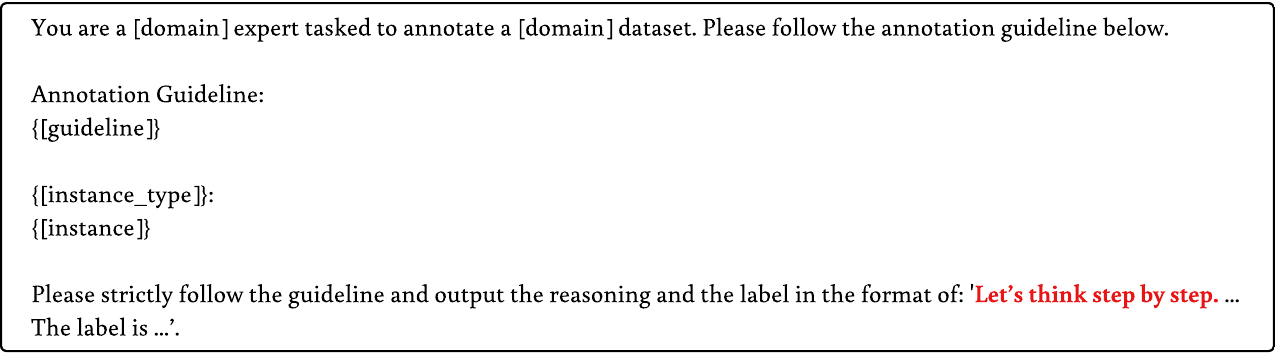}
  \caption{Multi-agent peer-discussion prompt template. Step 1: Generate initial annotation.}
  \label{fig:template-multi-agent-1}
\end{figure*}

\begin{figure*}[t]
  \centering
  \includegraphics[width=\linewidth]{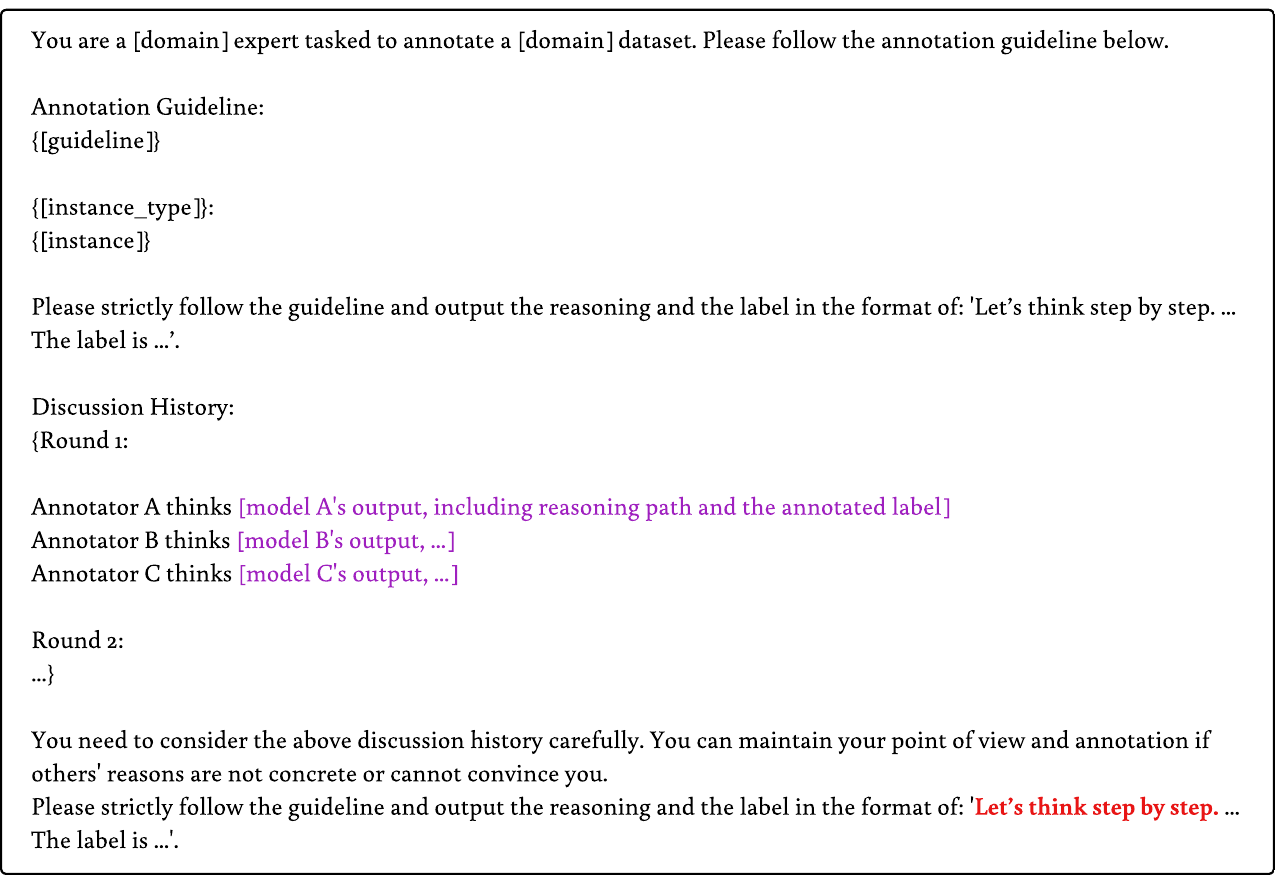}
  \caption{Multi-agent peer-discussion prompt template. Step 2: Discuss and re-annotate.}
  \label{fig:template-multi-agent-2}
\end{figure*}

We provide prompt templates of each methods from Figure~\ref{fig:template-vanilla} to Figure~\ref{fig:template-multi-agent-2}.

\begin{table*}[t]
  \small
  \centering
  \addtolength{\tabcolsep}{1pt}
  \resizebox{0.9\linewidth}{!}{
    \begin{tabular}{lllrr}
    \toprule
    \textbf{Domain} & \textbf{Dataset} & \textbf{Instance Type} & \textbf{$\#$Instances} & \textbf{$\#$Labels} \\
    \midrule
    \multirow{2}[2]{*}{Finance} & REFinD~\citep{kaur2023refind} & Sentence & 200 & 22 \\
          & FOMC~\citep{shah2023trillion}  & Sentence & 200 & 3 \\
    \midrule
    \multirow{2}[2]{*}{Law} & CUAD~\citep{hendrycks2021cuad}  & Clause & 200 & 32 \\
          & FoDS~\citep{guha2024legalbench}  & Excerpt & 200 & 7 \\
    \midrule
    Biomedicine & CODA-19~\citep{huang2020coda} & Paper Abstract & 200 & 5 \\
    \bottomrule
    \end{tabular}%
    }
    \caption{The statistics of existing domain-specific datasets we used in this paper.}
    \label{tab:dataset-statistic}
\end{table*}%

\section{Additional Results}\label{appendix-test}
We provide the accuracy and corresponding p-value of the comparison between reasoning models and non-reasoning models in Table~\ref{tab:appendix-acc} and Table~\ref{tab:appendix-pvalue}.

\begin{table}[htbp]
  \small    
  \centering
    \begin{tabular}{lrrrrrcr}
    \toprule
    \multicolumn{1}{c}{\multirow{2}[4]{*}{\textbf{Accuracy}}} & \multicolumn{2}{c}{\textbf{Finance}} &       & \multicolumn{2}{c}{\textbf{Law}} &       & \multicolumn{1}{l}{\textbf{Biomedicine}} \\
\cmidrule{2-3}\cmidrule{5-6}\cmidrule{8-8}          & \multicolumn{1}{c}{\textbf{REFinD}} & \multicolumn{1}{c}{\textbf{FOMC}} &       & \multicolumn{1}{c}{\textbf{CUAD}} & \multicolumn{1}{c}{\textbf{FoDS}} &       & \multicolumn{1}{c}{\textbf{CODA-19}} \\
    \midrule
    Best Vanilla & 67.5  & 71.5  &       & 86.5  & \textbf{47.0} &       & 79.5 \\
    Best CoT & 67.0  & 69.5  &       & 84.0  & 45.5  &       & 77.5 \\
    o3-mini (medium) & 71.5  & 73.0  &       & \textbf{87.0} & 46.5  &       & 82.0 \\
    Claude 3.7 Sonnet (thinking) & \textbf{73.0} & \textbf{74.0} &       & 86.5  & 46.5  &       & \textbf{84.5} \\
    \bottomrule
    \end{tabular}%
  \caption{Accuracy comparison between reasoning models and non-reasoning models. Text in \textbf{bold} indicates the highest accuracy for each dataset.}
  \label{tab:appendix-acc}%
\end{table}%

\begin{table}[htbp]
  \small
  \centering
    \begin{tabular}{lrrrrrcr}
    \toprule
    \multicolumn{1}{c}{\multirow{2}[4]{*}{\textbf{P-value}}} & \multicolumn{2}{c}{\textbf{Finance}} &       & \multicolumn{2}{c}{\textbf{Law}} &       & \multicolumn{1}{l}{\textbf{Biomedicine}} \\
\cmidrule{2-3}\cmidrule{5-6}\cmidrule{8-8}          & \multicolumn{1}{c}{\textbf{REFinD}} & \multicolumn{1}{c}{\textbf{FOMC}} &       & \multicolumn{1}{c}{\textbf{CUAD}} & \multicolumn{1}{c}{\textbf{FoDS}} &       & \multicolumn{1}{c}{\textbf{CODA-19}} \\
    \midrule
    o3-mini vs. Best vanilla & 0.230 & 0.728 &       & 1.000 & 1.000 &       & 0.533 \\
    o3-mini vs. Best CoT & 0.093 & 0.249 &       & 0.263 & 0.856 &       & 0.150 \\
    Claude 3.7 Sonnet vs. Best vanilla & 0.135 & 0.500 &       & 1.000 & 1.000 &       & 0.110 \\
    Claude 3.7 Sonnet vs. Best CoT & \textbf{0.043} & 0.163 &       & 0.332 & 0.839 &       & \textbf{0.013} \\
    \bottomrule
    \end{tabular}%
  \caption{P-values of the significance tests between reasoning models and non-reasoning models. P-values are rounded to three decimal places. Text in \textbf{bold} indicates the reasoning model is statistically significantly better than the best non-reasoning models, with p-value~$<$~0.05.}
  \label{tab:appendix-pvalue}%
\end{table}%

\end{document}